\documentclass[twoside]{article}

\usepackage[accepted]{aistats2026}
\usepackage[round]{natbib}

\usepackage[utf8]{inputenc}
\usepackage[T1]{fontenc}    
\usepackage{hyperref}      
\usepackage{url}           
\usepackage{booktabs}      
\usepackage{amsfonts}       
\usepackage{nicefrac}      
\usepackage{microtype}    
\usepackage{xcolor}        

\usepackage[textsize=tiny]{todonotes}
\usepackage{amsmath}

\usepackage{caption} 
\usepackage{subcaption}
\usepackage{wrapfig}
\usepackage{multirow}
\usepackage{multicol}
\usepackage[capitalize, nameinlink]{cleveref}
\usepackage{amsthm}

\theoremstyle{definition}
\newtheorem{definition}{Definition}
\newtheorem{observation}{Observation}

\crefname{equation}{Eq.}{Eqs.}
\crefname{figure}{Fig.}{Figs.}
\crefname{table}{Tab.}{Tabs.}
\crefname{eq}{Eq.}{Eqs.}
\crefname{condition}{Cond.}{Conds.}
\crefname{section}{Sec.}{Secs.}
\crefname{appendix}{App.}{Apps.}
\crefname{observation}{Obs.}{Obs.}

\hypersetup{
    colorlinks,
    linkcolor={red!50!black},
    citecolor={blue!50!black},
    urlcolor={blue!80!black}
}

\newcommand{\X}{\mathbf{X}}
\newcommand{\x}{\mathbf{x}}
\newcommand{\Y}{\mathbf{Y}}

\newcommand{\w}{\mathbf{w}}
\newcommand{\W}{\mathbf{W}}
\newcommand{\TP}{\boldsymbol{\Phi}}

\newcommand{\E}{\mathbb{E}}

\newcommand{\Var}{\text{Var}}

\newcommand{\cov}{\text{cov}}
\newcommand{\midxa}{\boldsymbol{\alpha}}
\newcommand{\midxb}{\boldsymbol{\beta}}

\begin{document}

%

%
\runningauthor{Johannes Exenberger, Sascha Ranftl, Robert Peharz}

\twocolumn[

\aistatstitle{Deep Polynomial Chaos Expansion}

\aistatsauthor{ 
  Johannes Exenberger 
  \And Sascha Ranftl$^{*}$ 
  \And  Robert Peharz$^{*}$ }

\aistatsaddress{ 
  CAIML, TU Wien \\ TU Graz
  \And  Brown University \\ Courant Institute NYU
  \And TU Graz \\ Graz Center for Machine Learning } ]

\begin{abstract}
Polynomial chaos expansion (PCE) is a classical and widely used surrogate modeling technique in physical simulation and uncertainty quantification.  
By taking a linear combination of a set of basis polynomials---orthonormal with respect to the distribution of uncertain input parameters---PCE enables tractable inference of key statistical quantities such as (conditional) means, variances, covariances, and Sobol sensitivity indices, which are essential for understanding the modeled system and identifying influential parameters and their interactions.  
The applicability of PCE to high-dimensional problems is limited by poor scalability, as the number of basis functions grows exponentially with the number of parameters.  
In this paper, we address this challenge by combining PCE with ideas from tractable probabilistic circuits, resulting in \emph{deep polynomial chaos expansion} (DeepPCE)---a deep generalization of PCE that scales effectively to high-dimensional input spaces.  
DeepPCE achieves predictive performance comparable to that of multilayer perceptrons (MLPs), while retaining PCE's ability to compute \emph{exact} statistical inferences via simple forward passes.  
In contrast, such computations in MLPs require costly and often inaccurate approximations, such as Monte Carlo integration.
\end{abstract}

We provide the code for DeepPCE at \url{https://github.com/jxnb/deep-polynomial-chaos-expansion}.

\section{INTRODUCTION}
\label{sec:Introduction}

\footnotetext{$^{*}$ Shared last authorship.}

Numerical simulations of partial differential equations (PDEs) are key tools in thermodynamics, solid-state physics, fluid dynamics, and many other areas of the physical sciences.
The accuracy of PDE solvers typically depends on several uncertain input parameters, such as boundary conditions, initial states, and physical properties.
Bayesian methods are often used to estimate these parameters, but they become infeasible with increasing input dimension or computationally demanding simulations.
A common workaround is to train a \emph{surrogate model} on a limited number of simulations, enabling efficient exploration of the parameter space.

\emph{Polynomial chaos expansion} (PCE) is a widely used surrogate model, originally introduced by \citet{wiener1938Homogeneous} for Gaussian inputs and later extended to a broader range of distributions \citep{xiu2002WienerAskey}.
The key idea is to approximate the \emph{response function}---i.e., the simulation output---as a linear combination of polynomial basis functions that are orthonormal with respect to the input distribution.
This orthonormality offers both theoretical and computational advantages.
In theory, PCE yields an optimal polynomial approximation in least-squares sense and, in the context of stochastic differential equations, allows coefficient computation similar to Galerkin projection \citep{galerkin1915electrical,ghanem1991Stochastic}.
In practice, PCE enhances data efficiency and simplifies the computation of (conditional) expectations, variances, covariances, and Sobol sensitivity indices \citep{sobol1990Sensitivity}.
These quantities have closed-form expressions when derived from PCE surrogates \citep{crestaux2009Polynomial,sudret2008Global}.

However, PCE scales poorly to high-dimensional parameter spaces, as the number of polynomial terms grows combinatorially with the number of inputs.
Truncation schemes \citep{muhlpfordt2018Comments}, sparse methods \citep{luthen2021Sparse}, and adaptive or greedy approaches \citep{blatman2011Adaptive,luthen2022Automatic} improve scalability to a certain extent, but ultimately break down in high-dimensional settings.
Due to these limitations, neural networks are increasingly used as surrogate models \citep{karniadakis2021Physicsinformed,raissi2019Physicsinformed}, as they often achieve better data fit for complex, high-dimensional physical systems.
A major limitation of neural surrogates is that expectations and Sobol indices are intractable and must be estimated via expensive and often imprecise Monte Carlo simulations.

In this paper, we propose a scalable and principled extension of PCE to high-dimensional inputs by combining it with a deep circuit architecture.
This idea is inspired by the framework of \emph{probabilistic circuits} \citep{choi2020Probabilistic}, which generalize shallow mixture models such as Gaussian mixture models (GMMs) into deep and structured representations.
Like PCE, GMMs and other mixture models suffer from the curse of dimensionality.
Deep probabilistic circuits overcome this problem by \emph{compactly representing exponentially many mixture components through circuit depth,} enabling effective modeling of high-dimensional distributions \citep{peharz2020Einsum,peharz2020Random}.
We apply the same principle to introduce \emph{deep polynomial chaos expansion} (DeepPCE), which can represent exponentially many orthogonal polynomial terms compactly.

In experiments, we show that DeepPCE scales PCE to problems with thousands of input parameters, far exceeding the capabilities of existing approaches.
In terms of data fit, DeepPCE performs comparably to other domain-agnostic neural network classes like multilayer perceptrons (MLPs).
Importantly, we derive exact and efficient methods for computing (conditional) means, variances, and covariances in DeepPCE, from which exact Sobol indices follow.
On benchmarks with known ground truth Sobol indices, DeepPCE matches these faithfully, whereas neural surrogates approximate them coarsely, even when using orders of magnitude more compute for Monte Carlo estimation.
In summary, our main contributions are:

\begin{itemize}
\item We propose a deep generalization of classical PCE, the \emph{deep polynomial chaos expansion} (DeepPCE), based on a deep circuit structure over orthogonal polynomial basis functions.
Unlike classical PCE, DeepPCE compactly represents a combinatorial number of polynomial terms and scales to high-dimensional parameter spaces.

\item We derive exact formulas for statistical moments (means, variances, covariances) and Sobol sensitivity indices within the DeepPCE framework. 

\item In experiments, we show that DeepPCE (i) matches the predictive performance of MLPs on high-dimensional regression tasks, (ii) enables scalable sensitivity analysis on common synthetic uncertainty quantification benchmarks and (iii) serves as a scalable surrogate for high-dimensional PDE benchmarks, namely Darcy flow and steady-state diffusion.
\end{itemize}

\section{BACKGROUND}

Random variables are written as uppercase letters (e.g.~$X$, $Y$, $Z$) and corresponding values with lowercase letters (e.g.~$x$, $y$, $z$). 
The set of integers $\{0, \dots, K\}$ is denoted by $[K]$.

\subsection{Polynomial Chaos Expansion (PCE)}
\label{sec:polynomial_chaos}

Numerical methods for modeling physical systems are often governed by a large number of uncertain input parameters, such as boundary and initial conditions or material properties. 
We denote these parameters by $\X = \{X_1, X_2, \dots, X_D\}$, forming a $D$-dimensional random vector. 
In PCE, it is common to assume a factorized input distribution $p(\x) = \prod_d p(x_d)$, although this assumption can be relaxed \citep{navarro2014Polynomial}.
The \emph{response function} $f^*(\X)$, i.e., the output of the numerical simulation, is approximated using a polynomial expansion.

For a single input variable $X = X_1$, the PCE of order $K$ is given as the linear combination
\begin{equation}
f_{\text{PCE}}(x) = \sum_{i=0}^K w_i \, \psi_i(x),
\end{equation}
where $\psi_i(x)$ are polynomial basis functions of degree $i$ and $w_i$ are real-valued coefficients. 
Crucially, the polynomials are chosen to be orthonormal with respect to the \emph{inner product induced by the distribution of $X$:}
\begin{equation}
\label{eq:orthonormality}
\E_{X} [\psi_i(X) \, \psi_j(X)] = \int p(x) \, \psi_i(x) \, \psi_j(x) \,\mathrm{d}x = 
\delta_{i,j},
\end{equation}
where $\delta_{i,j}$ is the Kronecker delta.
The original PCE formulation by \citet{wiener1938Homogeneous} assumed $X$ to be Gaussian, in which case the polynomials $\{\psi_i\}_{i=0}^K$ are Hermite polynomials.
\citet{xiu2002WienerAskey} extended this framework using the Askey scheme of hypergeometric polynomials, deriving orthonormal polynomials for a wide range of input distributions, such as Legendre polynomials for uniform distributions, Jacobi polynomials for Beta distributions, Krawtchouk polynomials for binomial distributions, etc.

For multiple inputs $\X = \{X_1, \dots, X_D\}$, multivariate basis functions are constructed as tensor products of univariate, orthonormal polynomials
\begin{equation}  
\label{eq:tensor_product_polynomials}
\TP_{\midxa}(\x) = \prod_{d = 1}^D \psi_{\alpha_d}(x_d),
\end{equation}
yielding the multivariate expansion
\begin{equation}
\label{eq:pce}
f_{\text{PCE}}(\x) = \sum_{\midxa \in \mathcal{A}} w_{\midxa} \TP_{\midxa}(\x).
\end{equation}
Here, $\midxa = (\alpha_1, \dots, \alpha_D) \in \mathcal{A}$ is a multi-index, where each $\alpha_d \in [K]$ denotes the polynomial degree associated with $X_d$, and $\mathcal{A}$ is the set of multi-indices for all terms in the expansion.
Because the input distribution $p(\x)$ is assumed to be factorized, and the univariate polynomials $\psi_{\alpha_d}$ in \cref{eq:tensor_product_polynomials} are orthonormal for each $X_d$, the tensor product polynomials are also orthonormal:
\begin{align}
\label{eq:orthonormal_tensor_products}
\E_{\X} [\TP_{\midxa}(\X) &\, \TP_{\midxa'}(\X)] = \nonumber \\
&= \int p(\x) \, \TP_{\midxa}(\x) \, \TP_{\midxa'}(\x) \,\mathrm{d}{\x} = \delta_{\midxa,\midxa'}.
\end{align}
The number of possible basis functions is bounded by
$
|\mathcal{A}| \leq \frac{(K + D)!}{K! D!},
$
which grows combinatorially with the number of inputs $D$ and the polynomial order $K$.
As this number quickly becomes infeasible with increasing dimensionality, truncation schemes are often applied \citep{sudret2015Polynomial}.
A common approach is to include only multi-indices satisfying $\| \midxa \|_p \leq t$, where $\|\cdot\|_p$ denotes the $p$-(quasi)norm for $0 < p \leq 1$, and $t$ is a threshold, usually set to the maximum polynomial order $K$. 
Truncation schemes or sparse adaptive methods \citep{luthen2021Sparse,luthen2022Automatic} improve the scalability of PCE only to a certain extent and require assumptions on the rank and sparsity of the basis sets.
Note that here, for simplicity, we describe the case of a scalar response function.
The case for multi-valued response functions is straightforward, by devising a PCE for each output.

\subsection{Sensitivity Analysis}   
\label{sec:sensitivity_analysis}

PCE has closed-form solutions for the expectation and variance of $f$.
Since the zeroth-order polynomial term, corresponding to $\midxa_0 = (0, \dots, 0)$, is a constant, it follows from \cref{eq:orthonormal_tensor_products} that $\E[\TP_{\midxa}(\X)]=0, \forall \midxa \neq \midxa_0$, hence the expectation is 
\begin{equation}
\label{eq:pce_expectation}
\E[f(\X)] = w_{\midxa_0}.
\end{equation}
Moreover, again due to \cref{eq:orthonormal_tensor_products}, all multiplicative cross-terms cancel in expectation and we get 
\begin{equation}
\label{eq:pce_var}
\Var(f(\X)) = \E[f_{\text{PCE}}^2(\X)]  - w_{\midxa_0}^2 = \sum_{\midxa \in  \mathcal{A} \setminus \{\midxa_0\}} w^2_{\midxa}.
\end{equation}
PCE also allows to compute ``advanced'' statistical quantities, such as variances of conditional expectations.
Let $\mathcal{I} \subseteq \{1, \dots, D\}$ be an index set for some parameters in $\X$ and $\lnot \mathcal{I} = \{1,\dots,D\} \setminus \mathcal{I}$.
It can be shown that

\begin{equation}
\label{eq:pce_conditional_variance}
\Var_{\X_{\mathcal{I}}}(\E_{\X_{\lnot \mathcal{I}}}[f(\X) \mid \X_{\mathcal{I}}])
= \sum_{\midxa \in  \mathcal{A}_{\mathcal{I}}} w^2_{\midxa}, 
\end{equation}

where $\mathcal{A}_{\mathcal{I}} = \{\midxa \colon \forall j \in \lnot\mathcal{I} \colon  \alpha_{j} = 0\}$.
These variances of conditional expectations allow to compute prominent sensitivity measures such as \emph{Sobol indices} \citep{sobol1990Sensitivity}.
These indices are normalized variances of the so-called \emph{Sobol decomposition}, which describes $f$ as a superposition of all possible $2^D$ interaction terms \citep{saltelli2008global}, including a constant, $D$ individual terms, $\binom{D}{2}$ pairwise terms, etc.

The \emph{first order Sobol indices} $S_{i}$ describe the direct contribution of each parameter $X_i$ to the variance of the response function, excluding contributions of $X_i$ via all interactive terms with other parameters:
\begin{equation}\label{eq:def-first-order-sobol}
  S_{i} = \frac{\Var_{X_{i}}(\E_{\lnot \X_{i}}[f(\X) \mid X_i])}{\Var(f(\X))}.
\end{equation}
The \emph{total effect Sobol indices} $S_{T_i}$ include also the variance contributions of $X_i$ caused by all interaction terms:
\begin{align}\label{eq:def-total-order-sobol}
  S_{T_i} &= \frac{\E_{\lnot \X_{i}}[\Var_{X_{i}}(f(\X) \mid \lnot \X_i)]}{\Var(f(\X))} \nonumber \\
          &= 1 - \frac{\Var_{\lnot \X_{i}}(\E_{X_{i}}[f(\X) \mid \lnot \X_i])}{\Var(f(\X))}.
\end{align}
In PCE, the indices are given via \cref{eq:pce_var,eq:pce_conditional_variance}. Other Sobol indices describing, for example, only pairwise interactions can also be computed in PCE.

\subsection{(Probabilistic) Circuits}
\label{sec:probabilisticcircuits}

Probabilistic circuits \citep{choi2020Probabilistic} are deep architectures composed of sum and product operations, whose structural constraints enable various tractable computations. 
Although primarily used for probabilistic modeling, they can be applied to regression tasks, in which case they are typically referred to as (structured) circuits.

\begin{definition}[(Probabilistic) Circuit]
\label{def:circuit}
A circuit $\mathcal{C} \equiv (\mathcal{G}, \W)$ is a computational graph $\mathcal{G}$ with parameters $\W$ defining a function $\mathcal{C}(\X)$ with input variables $\X = \{X_1, \dots, X_D\}$.
Here, $\mathcal{G}$ is an acyclic directed graph with three types of nodes: \emph{input nodes}, \emph{sum nodes} and \emph{product nodes}. 
In the following, let $\text{in}(c)$ be the set of input nodes of some node $c \in \mathcal{G}$.
An \emph{input} node $c$ (i.e.~$\text{ch}(c) = \emptyset$) represents a parameterized function $g_c(\X_c)$ over a subset of inputs $\X_c \subseteq \X$, called its \emph{scope}. 
Internal nodes (i.e.~$\text{in}(c) \not= \emptyset$) are either \emph{product} nodes or \emph{sum} nodes, representing operations $\prod_{c' \in \text{in}(c)} f_{c'}(\X_{c'})$ and $\sum_{c' \in \text{in}(c)} w_{c,c'}f_{c'}(\X_{c'})$, respectively, where $w_{c,c'}$ are the sum node parameters.
The scope of a sum or product node is recursively given as $\X_c = \bigcup_{c' \in \text{in}(c)}  \X_{c'}$.
A circuit is a \emph{probabilistic circuit} if $\mathcal{C}(\X)$ computes a probability distribution $p(\X)$ \citep{choi2020Probabilistic,vergari2021Compositional}.
\end{definition}

\begin{figure}[t]
    \centering
    \includegraphics{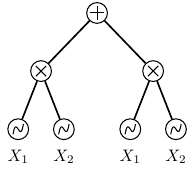}
        \caption{A simple \emph{smooth} and \emph{structured decomposable} circuit with input variables $\{X_1, X_2\}$.
         The input nodes encode tractable functions with inputs $X_1$ or $X_2$, followed by products over nodes with \emph{distinct} scopes $\{X_1\}$ or $\{X_2\}$. The root node computes a sum over product nodes with the same scope $\{X_1, X_2\}$.}
    \label{fig:circuit}
\end{figure}

A circuit is a special case of neural network, with non-linear functions in the input layer and linear units (sum nodes) and multiplicative units (product nodes) in subsequent layers. 
The set of parameters $\W$ is the union of all parameters of input functions and all sum weights.
Gradients of these are easily obtained by backpropagation.
Circuits enable tractable inference of different quantities through specific structural properties \citep{vergari2021Compositional} (\cref{fig:circuit}). A central structural constraint for our purpuse is \emph{(structured) decomposability}:

\begin{definition}[(Structured) Decomposability]
\label{def:structured_decomposability}
  A circuit is decomposable when the children of every product node have pairwise disjoint scope. 
  Formally, $\X_{c'} \cap \X_{c''} = \emptyset \; \forall c', c'' \in \text{in}(c) \; \forall c \in \Pi $, where $\Pi$ is the set of product nodes.
  A circuit is \emph{structured decomposable} if it is decomposable and if any pair of product nodes with the same scope decomposes the scope in the same way.
\end{definition}

Decomposability allows to rewrite high-dimensional integrals w.r.t.~inputs $\X$ as nested low-dimensional integrals by observing that (i) integration interchanges with summation and (ii) integration interchanges with decomposable multiplication \citep{peharz2015theoretical}. 
By induction, an integral over the circuit reduces to the integrals of the input nodes---each over the respective scope---followed by a single forward-pass.
Consequently, as long as integration over the input functions is tractable (e.g.~closed-form), the integral over the function represented by the whole circuit is tractable as well.

Often circuits are also assumed to be smooth \citep{choi2020Probabilistic}, which means that for each sum node, all input nodes have the same scope.
While not required for tractable integrations, smoothness yields well-defined \emph{probabilistic} circuits, i.e.~circuits that represent probability distributions, where tractable integrations allows efficient marginalization and conditioning. 
In this paper, we use circuits to establish a deep generalization of PCE that still allows to compute expectations, variances and Sobol indices exactly and efficiently.

\section{DEEP POLYNOMIAL CHAOS EXPANSION}
\label{sec:deep_pce}

\begin{figure}[t]
    \centering
    \includegraphics{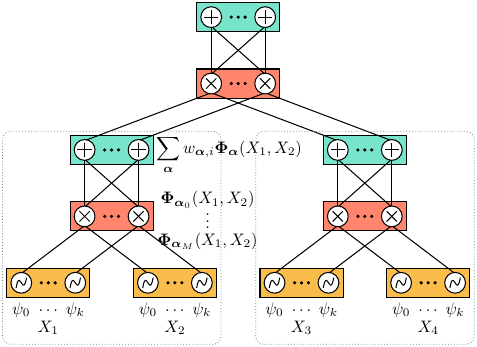}
        \caption{
        A DeepPCE over inputs $\{{X_1, X_2, X_3, X_4}\}$ with two input nodes, encoding PCEs over scopes $\{{X_1, X_2}\}$ and $\{{X_3, X_4}\}$.
        Inputs are expanded as orthogonal polynomials $\{\psi_i\}_{i=0}^K$. 
        A layer of outer-product nodes $\otimes$ forms the multivariate basis functions (tensor products) $\{\TP_{\midxa_0}, \dots, \TP_{\midxa_M} \}$. 
        A sum layer $\oplus$ with trainable network weights composes multiple PCEs over the same input scope but with distinct weights, overparameterizing the PCE layer for better expressivity.
        The input layer is followed by one or several blocks of Kronecker $\otimes$ or Hadamard products $\odot$ and weighted sums $\oplus$ over these products. Repeated sum-product layers can then be stacked to deeper structures. The depth of the DeepPCE is determined by the input dimensionality and the size of the input scopes.}
    \label{fig:deeppce}
\end{figure}

The core idea of DeepPCE is to use orthonormal PCE basis functions as input nodes in circuits, achieving a structured representation of a PCE that preserves the ability of PCEs to efficiently compute statistical moments and sensitivity indices at the input nodes while also allowing tractable inference based on circuit properties (\cref{fig:deeppce}). 
For tractable inference to work in circuit structures, it is required that the input nodes encode functions which have closed-form solutions; these functions do not necessarily have to be probability distributions. 
The integrals for the expectation $\E[f(\X)]$ and the variance $\Var({f(\X)})$ in PCEs can be computed efficiently from the expansion coefficients, as shown in \cref{eq:pce_expectation,eq:pce_var}.

From a computational perspective, DeepPCE is just as scalable as other circuits; constructing a PCE via hierarchical sums and products of PCEs with smaller scopes enables the modeling of high-dimensional functions beyond the reach of traditional PCE approaches, without any assumption on rank or sparsity of the polynomial basis.
This approach follows from two key observations: 

\begin{observation}[A PCE is a circuit]
Circuits can be interpreted as representations of polynomials \citep{choi2020Probabilistic,vergari2021Compositional}. Therefore, recalling \cref{eq:pce}, a PCE can be represented as a circuit, computing tensor products over polynomials $\TP_{\midxa}(\X) = \prod_{i \in \mathcal{I}} \psi_{\alpha_i}(X_i)$, where $\mathcal{I} \subseteq \{1, \dots, D \}$ are the indices of the variables in $\X$, followed by weighted sums over these products.
\end{observation}

\begin{observation}[A DeepPCE recovers a PCE] 
The distributive law of multiplication over addition allows to obtain a shallow representation $\mathcal{S}$ for any circuit $\mathcal{C}$ \citep{choi2020Probabilistic}. The shallow representation is

\begin{equation}
    \label{eq:circuit_shallow}
    \mathcal{S}(\X) = \sum_{i = 1}^P w_i \prod_{j = 1}^{M_i} g^{(i)}_{c_{j}}(\X^{(i)}_{c_{j}})
\end{equation}

where $P$ denotes the number of product nodes in the circuit and $M_i$ is the number of input nodes reachable from the $i$th product node, each encoding the function $g^{(i)}_{c_j}$ over scope $\X_{c_j}$. Recalling \cref{eq:tensor_product_polynomials} and \cref{eq:pce}, it becomes visible that a deep circuit with inputs encoding PCEs $( g_{c_1}(\X_{c_1}), \dots g_{c_N}(\X_{c_N}) )$ over scopes $\X_{c_n} \subseteq \X$ recovers a PCE over the whole scope $f_{\text{PCE}}(\X)$.
\end{observation}

\subsection{Architecture}

We implement the circuit as a layerwise tensorized computational graph \citep{loconte2025What}, following the structure of modern circuit implementations \citep{peharz2020Einsum,vergari2019Visualizing}.
We denote the multivariate circuit output as $\Y = \{Y_1, \dots, Y_O\}$, representing the function $\Y = f(\X)$, where the input nodes are defined as PCEs as described in \cref{sec:polynomial_chaos}.

\paragraph{PCE Layer}

Recall that the first layer of a circuit consists of multiple input nodes, each associated with a scope $\X_c \subseteq \X$.  
In contrast to probabilistic circuits, where input nodes encode probability distributions, each DeepPCE input node $c$ represents a PCE over scope $\X_c$, denoted as $g_c(\X_c)$ in the form of \cref{eq:pce}.  
We use overparameterized circuits \citep{loconte2025What}, with multiple input nodes encoding a PCE $g_{c,n}(\X_c)$ over the same scope, each with a distinct set of weights.  
In the special case where $\X_c = \X$, the entire input space is used as a single scope, recovering a shallow PCE.
We adopt a random partitioning strategy for assigning scopes to input nodes, randomly permuting the input vector $\X$ and splitting it into scopes of equal or similar length, which are then assigned to the input nodes.
This approach is inspired by \citet{peharz2020Random}; rather than mixing multiple randomly partitioned circuits at the root node, as proposed by the authors, we restrict the model to a single random partition of the input vector $\X$ to preserve circuit decomposability for tractable inference \citep{vergari2021Compositional}. 
Methods for learning the optimal structure of a circuit are actively developed \citep{adel2015Learning, gens2013Learning, yang2023Bayesian}, which could be integrated in the DeepPCE framework.

\paragraph{Sum-Product Layer}
Tractable circuits are composed of sequential sum and product nodes. 
Each product node computes the element-wise or outer product from its input vectors $\mathbf{u}, \mathbf{v}$ with length $M$ and scopes $\mathbf{X}_{c'}$ and $\mathbf{X}_{c''}$ respectively. In tensorized circuit representations, multiple such nodes are computed in parallel, either in Kronecker or Hadamard product layers \citep{loconte2025What}. 
For each product node $p_n$, there exist $L \geq 1$ individual sum nodes, each computing a sum over the same scope induced by the input node $p_n$ but with a distinct set of weights. The output of a sum node is $s_{n,l} = \sum_m^M w^{l, n}_{m} o_{m}$, where $w^{l, n}_{m}$ are the elements of the sum nodes' weight vector $\w_{l, n} \in \mathbf{W}_{L \times N \times M}$ and $\mathbf{W}$ is the complete weight tensor of the sum layer.
The number of sum nodes $L$ for each product node is a hyperparameter controlling the overparameterization of the circuit. 
At the output layer, $L = 1$ and each sum $s_n$ simply corresponds to one output dimension $Y_n \in \Y$.
The depth of the DeepPCE is determined by the input dimensionality $D$ and the size of the input scopes $\X_c$. 
In tensorized circuits, product and sum operations are usually abstracted in a single module for computational efficiency as shown in \citet{peharz2020Einsum}, yielding a sum-product layer.

\subsection{Training}
\label{sec:training}

Unlike classical PCE, DeepPCE does not admit a closed-form solution for the maximum likelihood estimate of the weights.  
Training therefore relies on standard gradient-based optimization.
Because PCEs are computed at the DeepPCE input layer, the values at the input nodes can span several orders of magnitude.
This is especially problematic for unbound input distributions such as the normal distribution, potentially destabilizing training.
This issue is further amplified by the multiplicative interactions in product nodes, which, in high-dimensional settings, often lead to vanishing or exploding outputs.  
As a result, DeepPCE may converge to poor local minima or fail to converge entirely.  

This challenge is not unique to DeepPCE; training circuit-based models in high-dimensional regimes is known to be difficult with existing optimizers \citep{liu2022Scaling}.
To mitigate the impact of large values introduced by higher-order polynomial terms, we draw inspiration from truncation strategies in the PCE literature \citep{blatman2011Adaptive}.  
Specifically, we initialize the weights of higher-order polynomial terms in the PCE input layer with lower variances than those of lower-order terms:
\begin{gather}
  \label{eq:weight_init_scaling}
  w_{\midxa} = x \cdot \lambda^{r} \\
  \text{with } r = \sum_{\alpha \in \midxa} \alpha, \quad x \sim \mathcal{N}(0, \sigma^2) \nonumber
\end{gather}
$\lambda \leq 1$ is a decay hyperparameter and $r$ is the rank of the polynomial term, given by the sum of the expansion terms' multi-index $\midxa$.
We find that this variance scaling improves both training stability and convergence, but does not mitigate the problem entirely. Robust parameter initialization strategies for the DeepPCE are thus an important part of further research. 

To further improve training stability, we apply batch normalization after each sum layer.  
Importantly, at test time, batch normalization simply amounts to a linear layer that can be absorbed into the constant (zeroth-degree) term of the polynomial expansion, preserving both orthogonality and tractability of the inference pass (see \cref{app:batch_norm}).
For the experiments presented in the paper, we perform randomized training of multiple models, dropping runs with bad parameter sets.

\subsection{Inference}

By building on both PCE and circuit properties, DeepPCE retains the ability to compute \emph{exact} statistical moments and sensitivity indices.
Due to the properties of PCEs, the operations reduce to simple masking operations of individual weights at the input nodes (the PCEs) of the DeepPCE, followed by a standard forward pass through the network. 
This way, we are able to compute expectations $\E[f(\X)]$, covariances $\cov(f(\X), f(\X))$, conditional expectations $\E[f(\X) \ | \ \x]$, conditional covariances $\cov(f(\X), f(\X) \ | \ \x)$, expectations of conditional covariances $\E[\cov(f(\X), f(\X) \ | \ \X_c \subset \X)]$ and covariances of conditional expectations $\cov(\E[f(\X) \ | \ \X_c \subset \X], \E[f(\X) \ | \ \X_c \subset \X])$. 
These quantities directly feed into the computation of Sobol indices for sensitivity analysis (see \cref{sec:sensitivity_analysis}).
As a sanity check, we compare all derived analytical moments with large-sample Monte Carlo estimates for evaluation. 
Results and detailed derivations of the inference pass are shown in \cref{app:inference}.

\paragraph{Expectations, covariances} 

As an example, we show how to compute the total model expectation $\E[f(\X)]$ and covariances $\cov(f(\X))$, making use of both PCE and circuit properties.
Under the assumption that $p(\X)$ factorizes, the expectation $\E[f(\X)]$ is represented by the same circuit as $f(\X)$, following from the fact that (i) sums commute with expectations, $\E[X + Y] = \E[X] + \E[Y]$, and (ii) factorized expectations distribute over decomposable products, $\E[XY] = \E[X] \E[Y] \text{ if } \, X \ {\perp \!\!\! \perp} \ Y $.
The single-dimensional expectations can hence be “pulled down” to the input nodes \citep{choi2020Probabilistic,peharz2015theoretical}.
At the input nodes, the expectation $\E[g(\X_c)]$ of a PCE node with scope $\X_c$ can be estimated directly from the weights as shown in \cref{eq:pce_expectation} by simply setting all weights $w_{\midxa_i} = 0 \ \forall i > 0$.
We get $\E[f(\X)]$ by using $\E[g(\X_c)]$ and performing a standard forward pass through the circuit.
Similarly, we get $\cov(f(\X),f(\X))$ by computing $\cov(g(\X_c),g(\X_c))$ at the input PCE---multiplicative cross-terms cancel due to orthogonality---according to \cref{eq:pce_var}, followed by a forward pass.
Applying the same rationale, we are able to compute other statistical quantities, such as conditional expectations, conditional covariances, expectations of conditional covariances and covariances of conditional expectations, as shown in \cref{app:inference}.

\section{RELATED WORK} 
\label{sec:related_work}

\paragraph{PCE} Numerous methods have been proposed to mitigate the curse of dimensionality in PCEs, aiming to extend their applicability to higher input dimensions and higher polynomial orders, including various (hyperbolic) truncation schemes \citep{muhlpfordt2018Comments}, compressed sensing \citep{mathelin2012Compressed}, variational inference \citep{bhattacharyya2020Global}, and sparse \citep{luthen2021Sparse}, adaptive \citep{blatman2011Adaptive, hampton2018Basis, jakeman2015Enhancing} or greedy \citep{doostan2011Nonadapted} approaches, and combinations thereof \citep{alemazkoor2017Divide, luthen2022Automatic}. 
Various approaches have explored integrating PCEs with machine learning \citep{torre2019Datadriven}, and their performance has been systematically compared in the literature \citep{kumar2025Comparative, shahane2019Uncertainty, shahzadi2021Deep}. 
Common strategies include (i) learning the PCE regression coefficients with (physics-informed) neural networks \citep{lutjens2021PCEPINNs, yao2023Deep, zheng2022Consistency}, (ii) learning the PCE basis functions with neural networks \citep{bahmani2025Neural, zhang2019Quantifying}, (iii) approximating the entire PCE as a neural network \citep{cooper2021Augmented, schwab2019Deep} or (iv) constructing the PCE on a learned, low-dimensional (latent) representations of input or output spaces \citep{kontolati2022Manifold, shustin2025PCENet}.
All these approaches aim at scalability at the expense of tractable inference.
\citet{oladyshkin2023Deep} proposed orthonormal polynomial bases as neural network activation functions, constructing a PCE in each neuron.
The basis functions are chosen to be orthonormal w.r.t. the previous layer's outputs using the arbitrary PCE approach \citep{oladyshkin2012Datadriven}, requiring re-orthonormalization after each weight update.
Unlike our DeepPCE framework, this method does not leverage structured, orthogonality-preserving hierarchical tensor products and does not recover a PCE.
Aside neural network-based approaches, PCEs have also been combined with bagging  \citep{elmocayd2021Datadriven} and  Gaussian process regression \citep{ranftl2021Bayesian,schobi2015Polynomialchaosbased}.
Lastly, PCEs can be extended to arbitrary probability measures \citep{oladyshkin2012Datadriven, wan2006Wiener}, dependent parameters \citep{jakeman2019Polynomial, navarro2014Polynomial, torre2019General}, discrete variables \citep{zheng2015Adaptive} and Bayesian PCEs \citep{arnst2010Identification, karagiannis2014Selection, ranftl2021Bayesian, tan2015Sequential}.

\paragraph{Probabilistic Circuits}
Orthogonality has been extensively studied in neural networks, particularly in weight matrices \citep{li2021Orthogonal}, latent spaces \citep{ha2021Adaptive}, and other settings \citep{achour2022Existence,mashhadi2021Parallel,yang1996Orthogonal}.
In contrast, orthonormal constructions within probabilistic circuits remain largely underexplored. 
\citet{loconte2025Faster} introduced orthonormal circuits that constrain input layers to orthonormal bases and additionally orthonormalize the circuit weights, which yields normalized squared circuits and faster marginalization.
Their notion of orthonormality is aiming at enabling faster probabilistic inferences, while DeepPCE leverages orthonormality and circuit properties to enable exact computations of statistical moments.
Earlier, \citet{adel2015Learning} employed SVD-based orthogonality for training circuits, but without constructing orthonormal structures.

\paragraph{Sensitivity Analysis and Uncertainty Quantification} 
Feature importance and sensitivity analysis play a critical role in uncertainty quantification, allowing to identify the most influential input parameters affecting a model's output. 
Proposed methods comprise model-agnostic, sampling-based methods, such as Shapley values \citep{ewald2024Guide,lundberg2017Unified}, and derivative-based \citep{lamboni2020Derivativebased,lamboni2021Multivariate,sobol2009Derivative} or variance-based methods \citep{homma1996Importance,lopiano2021Variancebased} such as Sobol sensitivity indices \citep{sobol1990Sensitivity}. 
Notably, the orthogonality of PCEs has been exploited to derive analytic expressions for Sobol sensitivity indices \citep{crestaux2009Polynomial,sudret2008Global}. 
Most generic machine learning models lack a built-in mechanism for variance decomposition, necessitating the use of expensive and potentially inaccurate Monte Carlo-based methods. 
Regarding uncertainty quantification \citet{abdar2021Review}, ensemble and Bayesian methods remain popular in the physical sciences \citep{psaros2023Uncertainty}. 
Unlike MLPs, probabilistic circuits like DeepPCE allow exact and scalable propagation of input uncertainties.

\begin{figure}
    \centering
    \includegraphics[width=\linewidth]{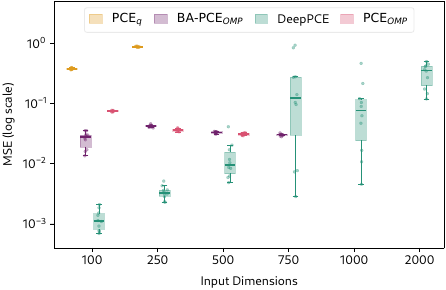}
        \caption{Test performances using the Bratley function with $D=[100, 250, 500, 750, 1000, 2000]$. 
        In contrast to classical PCE approaches, the DeepPCE manages to scale beyond 750 input dimensions and also outperforms the baseline PCEs in terms of predictive performance in lower dimensional settings.}
    \label{fig:bratley_scalability}
\end{figure}

\section{EXPERIMENTS}
\label{sec:experiments}

We evaluate the scalability and sensitivity analysis performance of the DeepPCE on several common benchmarks from the uncertainty quantification literature and show its performance as surrogate model on two high-dimensional PDE benchmarks compared to a UNet and a classic MLP.
The hyperparameters of the DeepPCE and other neural networks are tuned for each experiment.
Each experiment is performed 10 times for each model.
Additional experiments, comprehensive results of all runs and hyperparameter settings are described in \cref{app:experiments}.

\begin{figure*}[ht]
    \centering
    \begin{minipage}[t]{0.495\textwidth}
    \includegraphics[width=\linewidth]{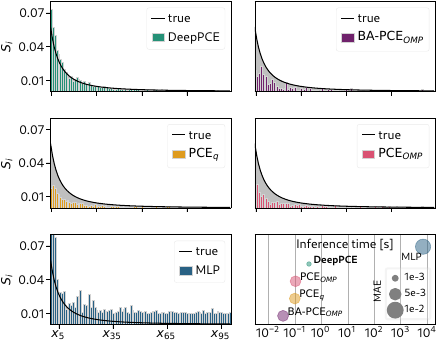}
    \end{minipage}
    \hfill
    \begin{minipage}[t]{0.495\textwidth}
    \includegraphics[width=\linewidth]{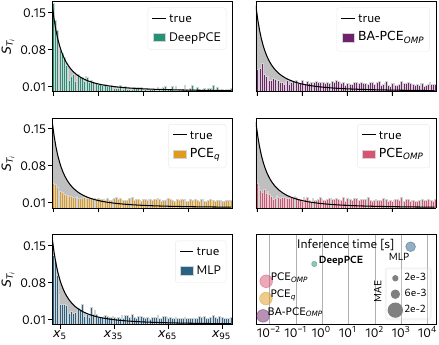}
    \end{minipage}
    \caption{First order Sobol indices (left) and total effect Sobol indices (right) for Sobol G* function with $D = 100$ (panels 1 - 5) and the measured wall clock time for an inference pass as well as mean average errors for the Sobol indices (panel 6). 
    The DeepPCE closely approximates to true variance contributions of each variable, outperforming all classical PCE variants. 
    The MLP's Sobol indices are obtained by Monte Carlo, using a total of $10^8$ samples to approximate the Sobol indices for each input $X_i$. 
    While it shows the worst performance of all models tested and fails to accurately represent the true variance contributions, the computation time to approximate the Sobol indices is longer by a factor of $10^3$ to $10^4$ compared to the DeepPCE (wall clock time).}
    \label{fig:sobol}
\end{figure*}

\begin{figure*}[ht]
    \centering
    \begin{minipage}[t]{0.47\textwidth}
      \includegraphics[width=\linewidth]{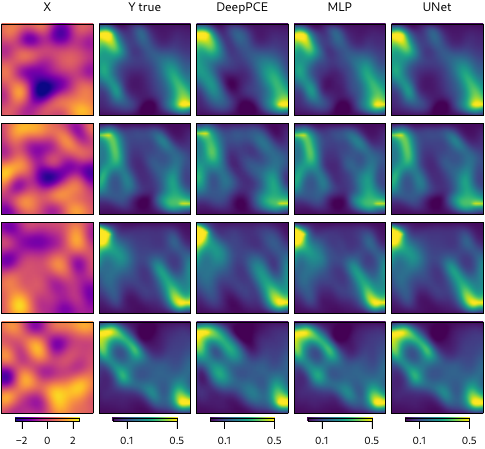}
    \end{minipage}
    \hfill
    \begin{minipage}[t]{0.47\textwidth}
      \includegraphics[width=\linewidth]{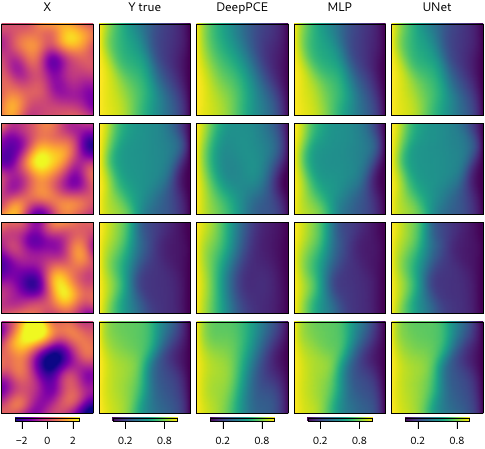}
    \end{minipage}
        \caption{Random test samples for the Darcy flow dataset (left) and the steady-state diffusion dataset (right) from DeepPCE, MLP and UNet. 
        The first column of panels show random input samples, the second column the corresponding true targets. 
        The results show that the DeepPCE manages to closely approximate the PDE results, offering predictive performance that is competitive against the UNet and the MLP, while also offering tractable and exact inference.}
    \label{fig:pdes}
\end{figure*}

\paragraph{PCE Benchmarks} 

We compare our DeepPCE with three classical PCEs using different approaches to generate the basis sets: (i) a PCE with a hyperbolic truncation scheme ($||\midxa||_{0.6} \leq K$) with $K = 5$ \citep{blatman2011Adaptive} ($\text{PCE}_\text{q}$), (ii) a sparse PCE based on orthogonal matching pursuit (OMP) ($\text{PCE}_{\text{OMP}}$) \citep{baptista2019Greedy}, initialized with the same basis as $\text{PCE}_q$, and (iii) a sparse PCE using OMP and the basis adaptive algorithm by \citet{jakeman2015Enhancing} ($\text{BA-PCE}_{\text{OMP}}$) to incrementally build a suitable basis. 
Various sparse and adaptive methods exist; we chose the sparse and adaptive variants based on their popularity and performance results from the comprehensive studies of \citet{luthen2021Sparse,luthen2022Automatic}, although they did not show a clear winner.
Based on the collection of test functions provided by \citet{wicaksono2023UQTestFuns}, we chose two common benchmark datasets for uncertainty quantification and sensitivity analysis that can be scaled to arbitrary dimensions: 
the \emph{Bratley Sum} function \citep{bratley1992Implementation,wicaksono2023UQTestFuns} and the \emph{Sobol G*} function \citep{saltelli2010Variance,wicaksono2023UQTestFuns}, which have known analyical expressions for Sobol sensitivity indices.
For each experiment, we use 10000 training samples and an additional 2000 samples for final evaluation.  
The training procedures for the DeepPCE and $\text{BA-PCE}_{\text{OMP}}$ demand a separate validation set, for which we used 2000 samples from the training set, keeping the total number of samples consistent for each model.
We provide detailed information on the function definitions and training results as well as additional experiments in \cref{app:experiments}.

\cref{fig:bratley_scalability} shows the performance of all benchmark PCEs and the DeepPCE on the Bratley Sum function for input dimensions $D = [100, 250, 500, 750, 1000, 2000]$. 
While the standard $\text{PCE}_q$ already breaks down with $D=250$, the sparse methods scale considerably better, with the $\text{BA-PCE}_{\text{OMP}}$ managing to scale up to 750 input dimensions. 
DeepPCE scales beyond this threshold and outperforms the sparse methods in terms of data fit in lower dimensional settings.

First order and total effect Sobol indices (\cref{eq:def-first-order-sobol,eq:def-total-order-sobol}) for the Sobol G* function for $D = 100$ are shown in \cref{fig:sobol}. 
The DeepPCE achieves the best approximation of both first order and total effect Sobol indices, outperforming all shallow PCE variants which have a tendency to underestimate the variances. 
For the MLP, we obtain Sobol indices by Monte Carlo sampling with $10^8$ samples.
As shown in \cref{fig:sobol}, the MLP does not manage to provide accurate estimates especially for first order Sobol indices.
Additionally, the Monte Carlo estimation has by far the longest runtime.
The wall clock time for computing MLP Sobol indices is longer by a factor of $10^3$ to $10^4$ compared to the analytical computation with the DeepPCE (hours vs. less than a second).

\paragraph{PDE datasets}

We use two PDE datasets to evaluate the capability of the DeepPCE as a surrogate model for high dimensional physics problems: 
(i) Darcy flow, a two-dimensional PDE and a common benchmark for data-driven physics models where the input is a permeability field following a log-normal distribution and the output is a velocity field, and (ii) a two-dimensional steady-state diffusion with a diffusion coefficient as input modeled by the logarithm of a Gaussian random field and a pressure field as output. 
In both datasets, the inputs are $64 \times 64$ dimensional fields, resulting in a data dimensionality of 4096---PCEs have so far not been applicable to such high dimensional problems. 
We use the Darcy flow dataset provided by \citet{zhu2018Bayesian}, the dataset for the steady-state diffusion PDE is based on \citet{tripathy2018Deep}.
For both datasets, we use 8000 training samples and 2000 validation samples. 
The test dataset for the Darcy flow PDE consists of 980 samples, while 2000 test samples are used for the steady state diffusion PDE.
Note that the input data in both PDEs is correlated; Sobol indices can thus not be inferred directly, yet the functions can still be approximated by the DeepPCE, allowing to evaluate its predictive performance in high dimensional settings.
Random test samples for both datasets are shown in \cref{fig:pdes}.
The UNet shows the best performance, due to its inductive bias specifically suited for image-type data, with relative MSEs in the range of $7 \cdot 10^{-5}$ for both PDE datasets. 
The average relative MSEs across all then runs for the MLP are $0.003$ for the Darcy flow experiment and for $0.0002$ for the steady state diffusion.
The DeepPCE shows average relative MSEs across all runs of $0.006$ for the Darcy flow experiment and $0.006$ for the steady state diffusion experiment.
While performing comparable to the MLP, the DeepPCE is the only model of the ones included in the comparsion that offers tractable and exact inference.

\section{CONCLUSION}

In this paper, we introduced Deep Polynomial Chaos Expansion (DeepPCE), a generalization of classical PCE that combines orthonormal polynomial bases within a deep, tractable circuit architecture.
DeepPCE overcomes the curse of dimensionality of PCE, without any assumptions on sparsity or on the order of interaction terms, while retaining PCE's ability to exactly compute statistical moments and Sobol sensitivity indices
In experiments, DeepPCE manages to accurately compute sensitivity indices and achieves predictive performance comparable to MLPs on high-dimensional PDE problems.

One limitation of DeepPCE is its sensitivity to the weight initialization, which can lead to convergence to bad local minima.
While our approach mitigates the problem to a certain extend, adressing parameter initialization and training techniques for circuits with polynomial basis functions is a promising direction for further research.
The reliance on factorized input distributions for sensitivity analysis suggests future extensions on correlated inputs. 
Extending the DeepPCE structure from simple random scopes by including, for example, structure learning methods is another relevant topic for further exploration. 
Lastly, our results suggest further exploration of orthogonality and (probabilistic) circuits, which remains largely underexplored.

\subsubsection*{Acknowledgements}
This research was funded in whole or in part by the Austrian Science Fund (FWF) 10.55776/COE12 and the project VENTUS (910263) under the FFG AI for Green program.

\bibliography{bibliography.bib}

@article{abdar2021Review,
  title = {A Review of Uncertainty Quantification in Deep Learning: {{Techniques}}, Applications and Challenges},
  shorttitle = {A Review of Uncertainty Quantification in Deep Learning},
  author = {Abdar, Moloud and Pourpanah, Farhad and Hussain, Sadiq and Rezazadegan, Dana and Liu, Li and Ghavamzadeh, Mohammad and Fieguth, Paul and Cao, Xiaochun and Khosravi, Abbas and Acharya, U. Rajendra and Makarenkov, Vladimir and Nahavandi, Saeid},
  year = 2021,
  journal = {Information Fusion},
  volume = {76},
  pages = {243--297}
}

@article{achour2022Existence,
  title = {Existence, {{Stability}} and {{Scalability}} of {{Orthogonal Convolutional Neural Networks}}},
  author = {Achour, El Mehdi and Malgouyres, Fran{\c c}ois and Mamalet, Franck},
  year = 2022,
  journal = {Journal of Machine Learning Research},
  volume = {23},
  number = {347},
  pages = {1--56}
}

@inproceedings{adel2015Learning,
  title = {Learning the {{Structure}} of {{Sum-Product Networks}} via an {{SVD-based Algorithm}}},
  booktitle = {Proceedings of the 31st {{Conference}} on {{Uncertainty}} in {{Artificial Intelligence}}},
  author = {Adel, Tameem and Balduzzi, David and Ghodsi, Ali},
  year = 2015
}

@article{alemazkoor2017Divide,
  title = {Divide and Conquer: {{An}} Incremental Sparsity Promoting Compressive Sampling Approach for Polynomial Chaos Expansions},
  shorttitle = {Divide and Conquer},
  author = {Alemazkoor, Negin and Meidani, Hadi},
  year = 2017,
  journal = {Computer Methods in Applied Mechanics and Engineering},
  volume = {318},
  pages = {937--956}
}

@article{arnst2010Identification,
  title = {Identification of {{Bayesian}} Posteriors for Coefficients of Chaos Expansions},
  author = {Arnst, M. and Ghanem, R. and Soize, C.},
  year = 2010,
  journal = {Journal of Computational Physics},
  volume = {229},
  number = {9},
  pages = {3134--3154}
}

@article{bahmani2025Neural,
  title = {Neural Chaos: {{A}} Spectral Stochastic Neural Operator},
  shorttitle = {Neural Chaos},
  author = {Bahmani, Bahador and Kevrekidis, Ioannis G. and Shields, Michael D.},
  year = 2025,
  journal = {Journal of Computational Physics},
  volume = {539},
  pages = {114233}
}

@article{baptista2019Greedy,
  title = {Some Greedy Algorithms for Sparse Polynomial Chaos Expansions},
  author = {Baptista, Ricardo and Stolbunov, Valentin and Nair, Prasanth B.},
  year = 2019,
  journal = {Journal of Computational Physics},
  volume = {387},
  pages = {303--325}
}

@article{bhattacharyya2020Global,
  title = {Global Sensitivity Analysis: {{A Bayesian}} Learning Based Polynomial Chaos Approach},
  shorttitle = {Global Sensitivity Analysis},
  author = {Bhattacharyya, Biswarup},
  year = 2020,
  journal = {Journal of Computational Physics},
  volume = {415},
  pages = {109539}
}

@article{blatman2011Adaptive,
  title = {Adaptive Sparse Polynomial Chaos Expansion Based on Least Angle Regression},
  author = {Blatman, G{\'e}raud and Sudret, Bruno},
  year = 2011,
  journal = {Journal of Computational Physics},
  volume = {230},
  number = {6},
  pages = {2345--2367}
}

@article{bratley1992Implementation,
  title = {Implementation and Tests of Low-Discrepancy Sequences},
  author = {Bratley, Paul and Fox, Bennett L. and Niederreiter, Harald},
  year = 1992,
  journal = {ACM Trans. Model. Comput. Simul.},
  volume = {2},
  number = {3},
  pages = {195--213}
}

@techreport{choi2020Probabilistic,
  title = {Probabilistic {{Circuits}}: {{A Unifying Framework}} for {{Tractable Probabilistic Models}}},
  author = {Choi, YooJung and Vergari, Antonio and {Van den Broeck}, Guy},
  year = 2020,
  address = {Los Angeles},
  institution = {University of California}
}

@phdthesis{cooper2021Augmented,
  title = {Augmented {{Neural Network Surrogate Models}} for {{Polynomial Chaos Expansions}} and {{Reduced Order Modeling}}},
  author = {Cooper, Rachel Gray},
  year = 2021,
  eprint = {10919/103423},
  eprinttype = {hdl},
  school = {Virginia Tech}
}

@article{crestaux2009Polynomial,
  title = {Polynomial Chaos Expansion for Sensitivity Analysis},
  author = {Crestaux, Thierry and Le Ma{\i}{\textasciicircum}tre, Olivier and Martinez, Jean-Marc},
  year = 2009,
  journal = {Reliability Engineering \& System Safety},
  series = {Special {{Issue}} on {{Sensitivity Analysis}}},
  volume = {94},
  number = {7},
  pages = {1161--1172}
}

@article{doostan2011Nonadapted,
  title = {A Non-Adapted Sparse Approximation of {{PDEs}} with Stochastic Inputs},
  author = {Doostan, Alireza and Owhadi, Houman},
  year = 2011,
  journal = {Journal of Computational Physics},
  volume = {230},
  number = {8},
  pages = {3015--3034}
}

@article{elmocayd2021Datadriven,
  title = {Data-Driven Polynomial Chaos Expansions for Characterization of Complex Fluid Rheology: {{Case}} Study of Phosphate Slurry},
  shorttitle = {Data-Driven Polynomial Chaos Expansions for Characterization of Complex Fluid Rheology},
  author = {El Mo{\c c}ayd, Nabil and Seaid, Mohammed},
  year = 2021,
  journal = {Reliability Engineering \& System Safety},
  volume = {216},
  pages = {107923}
}

@inproceedings{ewald2024Guide,
  title = {A {{Guide}} to~{{Feature Importance Methods}} for~{{Scientific Inference}}},
  booktitle = {Explainable {{Artificial Intelligence}}},
  author = {Ewald, Fiona Katharina and Bothmann, Ludwig and Wright, Marvin N. and Bischl, Bernd and Casalicchio, Giuseppe and K{\"o}nig, Gunnar},
  editor = {Longo, Luca and Lapuschkin, Sebastian and Seifert, Christin},
  year = 2024,
  pages = {440--464},
  publisher = {Springer Nature Switzerland},
  address = {Cham}
}

@article{galerkin1915electrical,
  title = {On Electrical Circuits for the Approximate Solution of the {{Laplace}} Equation},
  author = {Galerkin, {\relax BG}},
  year = 1915,
  journal = {Vestnik Inzh},
  volume = {19},
  pages = {897--908}
}

@inproceedings{gens2013Learning,
  title = {Learning the {{Structure}} of {{Sum-Product Networks}}},
  booktitle = {Proceedings of the 30th {{International Conference}} on {{Machine Learning}}},
  author = {Gens, Robert and Domingos, Pedro},
  year = 2013,
  pages = {873--880},
  publisher = {PMLR}
}

@book{ghanem1991Stochastic,
  title = {Stochastic {{Finite Elements}}: {{A Spectral Approach}}},
  shorttitle = {Stochastic {{Finite Elements}}},
  author = {Ghanem, Roger G. and Spanos, Pol D.},
  year = 1991,
  publisher = {Springer},
  address = {New York, NY}
}

@inproceedings{ha2021Adaptive,
  title = {Adaptive Wavelet Distillation from Neural Networks through Interpretations},
  booktitle = {Advances in {{Neural Information Processing Systems}}},
  author = {Ha, Wooseok and Singh, Chandan and Lanusse, Francois and Upadhyayula, Srigokul and Yu, Bin},
  year = 2021,
  volume = {34},
  pages = {20669--20682},
  publisher = {Curran Associates, Inc.}
}

@article{hampton2018Basis,
  title = {Basis Adaptive Sample Efficient Polynomial Chaos ({{BASE-PC}})},
  author = {Hampton, Jerrad and Doostan, Alireza},
  year = 2018,
  journal = {Journal of Computational Physics},
  volume = {371},
  pages = {20--49}
}

@article{homma1996Importance,
  title = {Importance Measures in Global Sensitivity Analysis of Nonlinear Models},
  author = {Homma, Toshimitsu and Saltelli, Andrea},
  year = 1996,
  journal = {Reliability Engineering \& System Safety},
  volume = {52},
  number = {1},
  pages = {1--17}
}

@article{jakeman2015Enhancing,
  title = {Enhancing {$\ell$}1-Minimization Estimates of Polynomial Chaos Expansions Using Basis Selection},
  author = {Jakeman, John D. and Eldred, M. S. and Sargsyan, K.},
  year = 2015,
  journal = {Journal of Computational Physics},
  volume = {289},
  pages = {18--34}
}

@article{jakeman2019Polynomial,
  title = {Polynomial Chaos Expansions for Dependent Random Variables},
  author = {Jakeman, John D. and Franzelin, Fabian and Narayan, Akil and Eldred, Michael and Plf{\"u}ger, Dirk},
  year = 2019,
  journal = {Computer Methods in Applied Mechanics and Engineering},
  volume = {351},
  pages = {643--666}
}

@article{karagiannis2014Selection,
  title = {Selection of Polynomial Chaos Bases via {{Bayesian}} Model Uncertainty Methods with Applications to Sparse Approximation of {{PDEs}} with Stochastic Inputs},
  author = {Karagiannis, Georgios and Lin, Guang},
  year = 2014,
  journal = {Journal of Computational Physics},
  volume = {259},
  pages = {114--134}
}

@article{karniadakis2021Physicsinformed,
  title = {Physics-Informed Machine Learning},
  author = {Karniadakis, George Em and Kevrekidis, Ioannis G. and Lu, Lu and Perdikaris, Paris and Wang, Sifan and Yang, Liu},
  year = 2021,
  journal = {Nature Reviews Physics},
  volume = {3},
  number = {6},
  pages = {422--440},
  publisher = {Nature Publishing Group}
}

@article{kontolati2022Manifold,
  title = {Manifold Learning-Based Polynomial Chaos Expansions for High-Dimensional Surrogate Models},
  author = {Kontolati, Katiana and Loukrezis, Dimitrios and dos Santos, Ketson R. M. and Giovanis, Dimitrios G. and Shields, Michael D.},
  year = 2022,
  journal = {International Journal for Uncertainty Quantification},
  volume = {12},
  number = {4},
  publisher = {Begel House Inc.}
}

@article{kucherenko2009Monte,
  title = {Monte {{Carlo}} Evaluation of Derivative-Based Global Sensitivity Measures},
  author = {Kucherenko, S. and {Rodriguez-Fernandez}, M. and Pantelides, C. and Shah, N.},
  year = 2009,
  journal = {Reliability Engineering \& System Safety},
  series = {Special {{Issue}} on {{Sensitivity Analysis}}},
  volume = {94},
  number = {7},
  pages = {1135--1148}
}

@article{kumar2025Comparative,
  title = {Comparative Analysis of {{Polynomial Chaos Expansion}} and {{Neural Networks}} for Inferring Elastic Modulus from Beam Deflection with {{Mat\'ern}} Covariance},
  author = {Kumar, Rakesh},
  year = 2025,
  journal = {Journal of Computational Science},
  volume = {91},
  pages = {102691}
}

@article{lamboni2020Derivativebased,
  title = {Derivative-Based Generalized Sensitivity Indices and {{Sobol}}' Indices},
  author = {Lamboni, Matieyendou},
  year = 2020,
  journal = {Mathematics and Computers in Simulation},
  volume = {170},
  pages = {236--256}
}

@article{lamboni2021Multivariate,
  title = {Multivariate Sensitivity Analysis and Derivative-Based Global Sensitivity Measures with Dependent Variables},
  author = {Lamboni, Matieyendou and Kucherenko, Sergei},
  year = 2021,
  journal = {Reliability Engineering \& System Safety},
  volume = {212},
  pages = {107519}
}

@article{li2021Orthogonal,
  title = {Orthogonal {{Deep Neural Networks}}},
  author = {Li, Shuai and Jia, Kui and Wen, Yuxin and Liu, Tongliang and Tao, Dacheng},
  year = 2021,
  journal = {IEEE Transactions on Pattern Analysis and Machine Intelligence},
  volume = {43},
  number = {4},
  pages = {1352--1368}
}

@inproceedings{liu2022Scaling,
  title = {Scaling {{Up Probabilistic Circuits}} by {{Latent Variable Distillation}}},
  booktitle = {The {{Eleventh International Conference}} on {{Learning Representations}} ({{ICLR}})},
  author = {Liu, Anji and Zhang, Honghua and den Broeck, Guy Van},
  year = 2022
}

@inproceedings{loconte2025Faster,
  title = {On {{Faster Marginalization}} with {{Squared Circuits}} via {{Orthonormalization}}},
  booktitle = {{{AAAI}} 25 {{Workshop}} on {{Connecting Lowrank Representations}} in {{AI}}},
  author = {Loconte, Lorenzo and Vergari, Antonio},
  year = 2025,
  eprint = {2412.07883},
  primaryclass = {cs},
  archiveprefix = {arXiv}
}

@article{loconte2025What,
  title = {What Is the {{Relationship}} between {{Tensor Factorizations}} and {{Circuits}} (and {{How Can We Exploit}} It)?},
  author = {Loconte, Lorenzo and Mari, Antonio and Gala, Gennaro and Peharz, Robert and de Campos, Cassio and Quaeghebeur, Erik and Vessio, Gennaro and Vergari, Antonio},
  year = 2025,
  journal = {Transactions on Machine Learning Research}
}

@article{lopiano2021Variancebased,
  title = {Variance-Based Sensitivity Analysis: {{The}} Quest for Better Estimators and Designs between Explorativity and Economy},
  shorttitle = {Variance-Based Sensitivity Analysis},
  author = {Lo Piano, Samuele and Ferretti, Federico and Puy, Arnald and Albrecht, Daniel and Saltelli, Andrea},
  year = 2021,
  journal = {Reliability Engineering \& System Safety},
  volume = {206},
  pages = {107300}
}

@inproceedings{lundberg2017Unified,
  title = {A {{Unified Approach}} to {{Interpreting Model Predictions}}},
  booktitle = {Advances in {{Neural Information Processing Systems}}},
  author = {Lundberg, Scott M and Lee, Su-In},
  year = 2017,
  volume = {30},
  publisher = {Curran Associates, Inc.}
}

@article{luthen2021Sparse,
  title = {Sparse {{Polynomial Chaos Expansions}}: {{Literature Survey}} and {{Benchmark}}},
  shorttitle = {Sparse {{Polynomial Chaos Expansions}}},
  author = {L{\"u}then, Nora and Marelli, Stefano and Sudret, Bruno},
  year = 2021,
  journal = {SIAM/ASA Journal on Uncertainty Quantification},
  volume = {9},
  number = {2},
  pages = {593--649},
  publisher = {{Society for Industrial and Applied Mathematics}}
}

@article{luthen2022Automatic,
  title = {Automatic Selection of Basis-Adaptive Sparse Polynomial Chaos Expansions for Engineering Applications},
  author = {L{\"u}then, Nora and Marelli, Stefano and Sudret, Bruno},
  year = 2022,
  journal = {International Journal for Uncertainty Quantification},
  volume = {12},
  number = {3},
  eprint = {2009.04800},
  primaryclass = {stat},
  archiveprefix = {arXiv}
}

@inproceedings{lutjens2021PCEPINNs,
  title = {{{PCE-PINNs}}: {{Physics-Informed Neural Networks}} for {{Uncertainty Propagation}} in {{Ocean Modeling}}},
  shorttitle = {{{PCE-PINNs}}},
  booktitle = {{{ICLR}} 2021 {{Workshop}} on {{AI}} for {{Modeling Oceans}} and {{Climate Change}}},
  author = {L{\"u}tjens, Bj{\"o}rn and Crawford, Catherine H. and Veillette, Mark and Newman, Dava},
  year = 2021,
  eprint = {2105.02939},
  primaryclass = {cs},
  archiveprefix = {arXiv}
}

@article{mashhadi2021Parallel,
  title = {Parallel Orthogonal Deep Neural Network},
  author = {Mashhadi, Peyman Sheikholharam and Nowaczyk, S{\l}awomir and Pashami, Sepideh},
  year = 2021,
  journal = {Neural Networks},
  volume = {140},
  pages = {167--183}
}

@article{mathelin2012Compressed,
  title = {A {{Compressed Sensing Approach}} for {{Partial Differential Equations}} with {{Random Input Data}}},
  author = {Mathelin, L. and Gallivan, K. A.},
  year = 2012,
  journal = {Communications in Computational Physics},
  volume = {12},
  number = {4},
  pages = {919--954}
}

@article{muhlpfordt2018Comments,
  title = {Comments on {{Truncation Errors}} for {{Polynomial Chaos Expansions}}},
  author = {M{\"u}hlpfordt, Tillmann and Findeisen, Rolf and Hagenmeyer, Veit and Faulwasser, Timm},
  year = 2018,
  journal = {IEEE Control Systems Letters},
  volume = {2},
  number = {1},
  pages = {169--174}
}

@misc{navarro2014Polynomial,
  title = {Polynomial {{Chaos Expansion}} for General Multivariate Distributions with Correlated Variables},
  author = {Navarro, Maria and Witteveen, Jeroen and Blom, Joke},
  year = 2014,
  number = {arXiv:1406.5483},
  eprint = {1406.5483},
  primaryclass = {math},
  publisher = {arXiv},
  archiveprefix = {arXiv}
}

@article{oladyshkin2012Datadriven,
  title = {Data-Driven Uncertainty Quantification Using the Arbitrary Polynomial Chaos Expansion},
  author = {Oladyshkin, S. and Nowak, W.},
  year = 2012,
  journal = {Reliability Engineering \& System Safety},
  volume = {106},
  pages = {179--190}
}

@article{oladyshkin2023Deep,
  title = {The Deep Arbitrary Polynomial Chaos Neural Network or How {{Deep Artificial Neural Networks}} Could Benefit from Data-Driven Homogeneous Chaos Theory},
  author = {Oladyshkin, Sergey and Praditia, Timothy and Kroeker, Ilja and Mohammadi, Farid and Nowak, Wolfgang and Otte, Sebastian},
  year = 2023,
  journal = {Neural Networks},
  volume = {166},
  pages = {85--104}
}

@inproceedings{peharz2015Theoretical,
  title = {On {{Theoretical Properties}} of {{Sum-Product Networks}}},
  booktitle = {Proceedings of the {{Eighteenth International Conference}} on {{Artificial Intelligence}} and {{Statistics}}},
  author = {Peharz, Robert and Tschiatschek, Sebastian and Pernkopf, Franz and Domingos, Pedro},
  year = 2015,
  pages = {744--752},
  publisher = {PMLR}
}

@inproceedings{peharz2020Einsum,
  title = {Einsum {{Networks}}: {{Fast}} and {{Scalable Learning}} of {{Tractable Probabilistic Circuits}}},
  shorttitle = {Einsum {{Networks}}},
  booktitle = {Proceedings of the 37th {{International Conference}} on {{Machine Learning}}},
  author = {Peharz, Robert and Lang, Steven and Vergari, Antonio and Stelzner, Karl and Molina, Alejandro and Trapp, Martin and Broeck, Guy Van Den and Kersting, Kristian and Ghahramani, Zoubin},
  year = 2020,
  pages = {7563--7574},
  publisher = {PMLR}
}

@inproceedings{peharz2020Random,
  title = {Random {{Sum-Product Networks}}: {{A Simple}} and {{Effective Approach}} to {{Probabilistic Deep Learning}}},
  shorttitle = {Random {{Sum-Product Networks}}},
  booktitle = {Proceedings of {{The}} 35th {{Uncertainty}} in {{Artificial Intelligence Conference}}},
  author = {Peharz, Robert and Vergari, Antonio and Stelzner, Karl and Molina, Alejandro and Shao, Xiaoting and Trapp, Martin and Kersting, Kristian and Ghahramani, Zoubin},
  year = 2020,
  pages = {334--344},
  publisher = {PMLR}
}

@article{psaros2023Uncertainty,
  title = {Uncertainty Quantification in Scientific Machine Learning: {{Methods}}, Metrics, and Comparisons},
  shorttitle = {Uncertainty Quantification in Scientific Machine Learning},
  author = {Psaros, Apostolos F. and Meng, Xuhui and Zou, Zongren and Guo, Ling and Karniadakis, George Em},
  year = 2023,
  journal = {Journal of Computational Physics},
  volume = {477},
  pages = {111902}
}

@article{raissi2019Physicsinformed,
  title = {Physics-Informed Neural Networks: {{A}} Deep Learning Framework for Solving Forward and Inverse Problems Involving Nonlinear Partial Differential Equations},
  shorttitle = {Physics-Informed Neural Networks},
  author = {Raissi, M. and Perdikaris, P. and Karniadakis, G. E.},
  year = 2019,
  journal = {Journal of Computational Physics},
  volume = {378},
  pages = {686--707}
}

@article{ranftl2021Bayesian,
  title = {Bayesian {{Surrogate Analysis}} and {{Uncertainty Propagation}}},
  author = {Ranftl, Sascha and {von der Linden}, Wolfgang},
  year = 2021,
  journal = {Physical Sciences Forum},
  volume = {3},
  number = {1},
  pages = {6},
  publisher = {Multidisciplinary Digital Publishing Institute}
}

@article{saltelli1995Use,
  title = {About the Use of Rank Transformation in Sensitivity Analysis of Model Output},
  author = {Saltelli, Andrea and Sobol', Ilya M},
  year = 1995,
  journal = {Reliability Engineering \& System Safety},
  volume = {50},
  number = {3},
  pages = {225--239}
}

@book{saltelli2008Global,
  title = {Global {{Sensitivity Analysis}}: {{The Primer}}},
  shorttitle = {Global {{Sensitivity Analysis}}},
  author = {Saltelli, Andrea and Ratto, Marco and Andres, Terry and Campolongo, Francesca and Cariboni, Jessica and Gatelli, Debora and Saisana, Michaela and Tarantola, Stefano},
  year = 2008,
  publisher = {John Wiley \& Sons},
  googlebooks = {wAssmt2vumgC}
}

@article{saltelli2010Variance,
  title = {Variance Based Sensitivity Analysis of Model Output. {{Design}} and Estimator for the Total Sensitivity Index},
  author = {Saltelli, Andrea and Annoni, Paola and Azzini, Ivano and Campolongo, Francesca and Ratto, Marco and Tarantola, Stefano},
  year = 2010,
  journal = {Computer Physics Communications},
  volume = {181},
  number = {2},
  pages = {259--270}
}

@article{schobi2015Polynomialchaosbased,
  title = {Polynomial-Chaos-Based Kriging},
  author = {Schobi, Roland and Sudret, Bruno and Wiart, Joe},
  year = 2015,
  journal = {International Journal for Uncertainty Quantification},
  volume = {5},
  number = {2},
  publisher = {Begel House Inc.}
}

@article{schwab2019Deep,
  title = {Deep Learning in High Dimension: {{Neural}} Network Expression Rates for Generalized Polynomial Chaos Expansions in {{UQ}}},
  shorttitle = {Deep Learning in High Dimension},
  author = {Schwab, Christoph and Zech, Jakob},
  year = 2019,
  journal = {Analysis and Applications},
  volume = {17},
  number = {01},
  pages = {19--55},
  publisher = {World Scientific Publishing Co.}
}

@article{shahane2019Uncertainty,
  title = {Uncertainty Quantification in Three Dimensional Natural Convection Using Polynomial Chaos Expansion and Deep Neural Networks},
  author = {Shahane, Shantanu and Aluru, Narayana R. and Vanka, Surya Pratap},
  year = 2019,
  journal = {International Journal of Heat and Mass Transfer},
  volume = {139},
  pages = {613--631}
}

@article{shahzadi2021Deep,
  title = {Deep {{Neural Network}} and {{Polynomial Chaos Expansion-Based Surrogate Models}} for {{Sensitivity}} and {{Uncertainty Propagation}}: {{An Application}} to a {{Rockfill Dam}}},
  shorttitle = {Deep {{Neural Network}} and {{Polynomial Chaos Expansion-Based Surrogate Models}} for {{Sensitivity}} and {{Uncertainty Propagation}}},
  author = {Shahzadi, Gullnaz and Soula{\"i}mani, Azzeddine},
  year = 2021,
  journal = {Water},
  volume = {13},
  number = {13},
  publisher = {publisher}
}

@misc{shustin2025PCENet,
  title = {{{PCENet}}: {{High Dimensional Surrogate Modeling}} for {{Learning Uncertainty}}},
  shorttitle = {{{PCENet}}},
  author = {Shustin, Paz Fink and Ubaru, Shashanka and Zimo{\'n}, Ma{\l}gorzata J. and Lu, Songtao and Kalantzis, Vasileios and Horesh, Lior and Avron, Haim},
  year = 2025,
  number = {arXiv:2202.05063},
  eprint = {2202.05063},
  primaryclass = {cs},
  publisher = {arXiv},
  archiveprefix = {arXiv}
}

@article{sobol1990sensitivity,
  title = {On Sensitivity Estimation for Nonlinear Mathematical Models},
  author = {Sobol', Il'ya Meerovich},
  year = 1990,
  journal = {Matematicheskoe modelirovanie},
  volume = {2},
  number = {1},
  pages = {112--118},
  publisher = {Russian Academy of Sciences, Branch of Mathematical Sciences}
}

@article{sobol2009Derivative,
  title = {Derivative Based Global Sensitivity Measures and Their Link with Global Sensitivity Indices},
  author = {Sobol', I. M. and Kucherenko, S.},
  year = 2009,
  journal = {Mathematics and Computers in Simulation},
  volume = {79},
  number = {10},
  pages = {3009--3017}
}

@article{sudret2008Global,
  title = {Global Sensitivity Analysis Using Polynomial Chaos Expansions},
  author = {Sudret, Bruno},
  year = 2008,
  journal = {Reliability Engineering \& System Safety},
  series = {Bayesian {{Networks}} in {{Dependability}}},
  volume = {93},
  number = {7},
  pages = {964--979}
}

@article{sudret2015Polynomial,
  title = {Polynomial Chaos Expansions and Stochastic Finite Element Methods},
  author = {Sudret, Bruno},
  year = 2015,
  journal = {Risk and Reliability in Geotechnical Engineering},
  pages = {265--300}
}

@article{tan2015Sequential,
  title = {Sequential {{Bayesian Polynomial Chaos Model Selection}} for {{Estimation}} of {{Sensitivity Indices}}},
  author = {Tan, Matthias Hwai Yong},
  year = 2015,
  journal = {SIAM/ASA Journal on Uncertainty Quantification},
  volume = {3},
  number = {1},
  pages = {146--168},
  publisher = {{Society for Industrial and Applied Mathematics}}
}

@article{torre2019Datadriven,
  title = {Data-Driven Polynomial Chaos Expansion for Machine Learning Regression},
  author = {Torre, Emiliano and Marelli, Stefano and Embrechts, Paul and Sudret, Bruno},
  year = 2019,
  journal = {Journal of Computational Physics},
  volume = {388},
  pages = {601--623}
}

@article{torre2019General,
  title = {A General Framework for Data-Driven Uncertainty Quantification under Complex Input Dependencies Using Vine Copulas},
  author = {Torre, Emiliano and Marelli, Stefano and Embrechts, Paul and Sudret, Bruno},
  year = 2019,
  journal = {Probabilistic Engineering Mechanics},
  volume = {55},
  pages = {1--16}
}

@article{tripathy2018Deep,
  title = {Deep {{UQ}}: {{Learning}} Deep Neural Network Surrogate Models for High Dimensional Uncertainty Quantification},
  shorttitle = {Deep {{UQ}}},
  author = {Tripathy, Rohit K. and Bilionis, Ilias},
  year = 2018,
  journal = {Journal of Computational Physics},
  volume = {375},
  pages = {565--588}
}

@article{vergari2019Visualizing,
  title = {Visualizing and Understanding {{Sum-Product Networks}}},
  author = {Vergari, Antonio and Di Mauro, Nicola and Esposito, Floriana},
  year = 2019,
  journal = {Machine Learning},
  volume = {108},
  number = {4},
  pages = {551--573}
}

@inproceedings{vergari2021Compositional,
  title = {A {{Compositional Atlas}} of {{Tractable Circuit Operations}} for {{Probabilistic Inference}}},
  booktitle = {Advances in {{Neural Information Processing Systems}}},
  author = {Vergari, Antonio and Choi, YooJung and Liu, Anji and Teso, Stefano and {Van den Broeck}, Guy},
  year = 2021,
  volume = {34},
  pages = {13189--13201},
  publisher = {Curran Associates, Inc.}
}

@article{wan2006Wiener,
  title = {Beyond {{Wiener}}--{{Askey Expansions}}: {{Handling Arbitrary PDFs}}},
  shorttitle = {Beyond {{Wiener}}--{{Askey Expansions}}},
  author = {Wan, Xiaoliang and Karniadakis, George Em},
  year = 2006,
  journal = {Journal of Scientific Computing},
  volume = {27},
  number = {1},
  pages = {455--464}
}

@article{wicaksono2023UQTestFuns,
  title = {{{UQTestFuns}}: {{A Python3}} Library of Uncertainty Quantification ({{UQ}}) Test Functions},
  shorttitle = {{{UQTestFuns}}},
  author = {Wicaksono, Damar and Hecht, Michael},
  year = 2023,
  journal = {Journal of Open Source Software},
  volume = {8},
  number = {90},
  pages = {5671}
}

@article{wiener1938Homogeneous,
  title = {The {{Homogeneous Chaos}}},
  author = {Wiener, Norbert},
  year = 1938,
  journal = {American Journal of Mathematics},
  volume = {60},
  number = {4},
  eprint = {2371268},
  eprinttype = {jstor},
  pages = {897--936},
  publisher = {Johns Hopkins University Press}
}

@article{xiu2002WienerAskey,
  title = {The {{Wiener-Askey Polynomial Chaos}} for {{Stochastic Differential Equations}}},
  author = {Xiu, Dongbin and Karniadakis, George Em},
  year = 2002,
  journal = {SIAM Journal on Scientific Computing},
  volume = {24},
  number = {2},
  pages = {619--644}
}

@article{yang1996Orthogonal,
  title = {An Orthogonal Neural Network for Function Approximation},
  author = {Yang, Shiow-Shung and Tseng, Ching-Shiow},
  year = 1996,
  journal = {IEEE Transactions on Systems, Man, and Cybernetics, Part B (Cybernetics)},
  volume = {26},
  number = {5},
  pages = {779--785}
}

@inproceedings{yang2023Bayesian,
  title = {Bayesian {{Structure Scores}} for {{Probabilistic Circuits}}},
  booktitle = {Proceedings of {{The}} 26th {{International Conference}} on {{Artificial Intelligence}} and {{Statistics}}},
  author = {Yang, Yang and Gala, Gennaro and Peharz, Robert},
  year = 2023,
  pages = {563--575},
  publisher = {PMLR}
}

@article{yao2023Deep,
  title = {Deep Adaptive Arbitrary Polynomial Chaos Expansion: {{A}} Mini-Data-Driven Semi-Supervised Method for Uncertainty Quantification},
  shorttitle = {Deep Adaptive Arbitrary Polynomial Chaos Expansion},
  author = {Yao, Wen and Zheng, Xiaohu and Zhang, Jun and Wang, Ning and Tang, Guijian},
  year = 2023,
  journal = {Reliability Engineering \& System Safety},
  volume = {229},
  pages = {108813}
}

@article{zhang2019Quantifying,
  title = {Quantifying Total Uncertainty in Physics-Informed Neural Networks for Solving Forward and Inverse Stochastic Problems},
  author = {Zhang, Dongkun and Lu, Lu and Guo, Ling and Karniadakis, George Em},
  year = 2019,
  journal = {Journal of Computational Physics},
  volume = {397},
  pages = {108850}
}

@article{zheng2015Adaptive,
  title = {Adaptive Multi-Element Polynomial Chaos with Discrete Measure: {{Algorithms}} and Application to {{SPDEs}}},
  shorttitle = {Adaptive Multi-Element Polynomial Chaos with Discrete Measure},
  author = {Zheng, Mengdi and Wan, Xiaoliang and Karniadakis, George Em},
  year = 2015,
  journal = {Applied Numerical Mathematics},
  volume = {90},
  pages = {91--110}
}

@article{zheng2022Consistency,
  title = {Consistency Regularization-Based Deep Polynomial Chaos Neural Network Method for Reliability Analysis},
  author = {Zheng, Xiaohu and Yao, Wen and Zhang, Yunyang and Zhang, Xiaoya},
  year = 2022,
  journal = {Reliability Engineering \& System Safety},
  volume = {227},
  pages = {108732}
}

@article{zhu2018Bayesian,
  title = {Bayesian Deep Convolutional Encoder--Decoder Networks for Surrogate Modeling and Uncertainty Quantification},
  author = {Zhu, Yinhao and Zabaras, Nicholas},
  year = 2018,
  journal = {Journal of Computational Physics},
  volume = {366},
  pages = {415--447}
}

\section*{Checklist}

\begin{enumerate}

  \item For all models and algorithms presented, check if you include:
  \begin{enumerate}
    \item A clear description of the mathematical setting, assumptions, algorithm, and/or model. [Yes] Detailed explanations of all models are in \cref{app:experiments}.
    \item An analysis of the properties and complexity (time, space, sample size) of any algorithm. [Not Applicable]
    \item (Optional) Anonymized source code, with specification of all dependencies, including external libraries. [Yes] We will publish the code in github after acceptance.
  \end{enumerate}

  \item For any theoretical claim, check if you include:
  \begin{enumerate}
    \item Statements of the full set of assumptions of all theoretical results. [Yes]
    \item Complete proofs of all theoretical results. [Yes] The theoretical derivation of the inference pass is shown in \cref{app:inference}
    \item Clear explanations of any assumptions. [Yes] We state all relevant assumptions in the paper, such as in Section 2.   
  \end{enumerate}

  \item For all figures and tables that present empirical results, check if you include:
  \begin{enumerate}
    \item The code, data, and instructions needed to reproduce the main experimental results (either in the supplemental material or as a URL). [Yes] The experiment scripts. including random seeds, are published with the source code on Github.
    \item All the training details (e.g., data splits, hyperparameters, how they were chosen). [Yes] See \cref{sec:experiments} and \cref{app:experiments}.
    \item A clear definition of the specific measure or statistics and error bars (e.g., with respect to the random seed after running experiments multiple times). [Yes]
    \item A description of the computing infrastructure used. (e.g., type of GPUs, internal cluster, or cloud provider). [Yes] See \cref{app:experiments}
  \end{enumerate}

  \item If you are using existing assets (e.g., code, data, models) or curating/releasing new assets, check if you include:
  \begin{enumerate}
    \item Citations of the creator If your work uses existing assets. [Yes]
    \item The license information of the assets, if applicable. [Not Applicable]
    \item New assets either in the supplemental material or as a URL, if applicable. [Not Applicable]
    \item Information about consent from data providers/curators. [Not Applicable]
    \item Discussion of sensible content if applicable, e.g., personally identifiable information or offensive content. [Not Applicable]
  \end{enumerate}

  \item If you used crowdsourcing or conducted research with human subjects, check if you include:
  \begin{enumerate}
    \item The full text of instructions given to participants and screenshots. [Not Applicable]
    \item Descriptions of potential participant risks, with links to Institutional Review Board (IRB) approvals if applicable. [Not Applicable]
    \item The estimated hourly wage paid to participants and the total amount spent on participant compensation. [Not Applicable]
  \end{enumerate}

\end{enumerate}

\clearpage
\appendix
\thispagestyle{empty}
\onecolumn
\aistatstitle{Supplementary Material}

\section{CIRCUITS}
\label{app:circuit_preliminaries}

The DeepPCE shares its structure with (probabilistic) circuits --- the constraints imposed on the structure of circuits allow the tractable computation of different inference queries \citep{vergari2021Compositional}. 
For detailed explanations and proofs of the theoretical foundations and necessary and sufficient conditions for tractability, we refer to \citet{choi2020Probabilistic, vergari2021Compositional, loconte2025What}. 
Here, we will provide a higher-level summary of the important concepts. 
Similar to other frameworks \citep{peharz2020Einsum, loconte2025What}, the DeepPCE is implemented as layerwise tensorized computational graph, parallelizing the operations inside the circuit. 
It still consists of the same basic computational \emph{nodes}, which we explain briefly. 
An inner node $c$ (a node that is not an input node) in the circuit receives inputs from one or more other circuit nodes, its \emph{inputs}, which we denote as $\text{in}(c)$. 
Each node encodes a function, denoted as $h_c$ over a subset of input variables $\X_c \subseteq \X$, also called the \emph{scope} of a node. 
A circuit has only three distinct computational nodes:

\paragraph{Input node} 

A \emph{input node} $c$ encodes an integrable parameterized function over input variables $\X_c \subseteq \X$. 
While a input node in probabilistic circuits encodes a probability distribution, in the case of DeepPCE it encodes a PCE, which we denote as $g$. 
Recalling \cref{sec:polynomial_chaos}, a PCE over scope $\X_c$ has the form

\begin{equation}
    g(\X_c) = \sum_{\midxa \in \mathcal{A}_c} w_{\midxa} \TP_{\midxa}(\X_c).
\end{equation}

\paragraph{Product node} 

A \emph{product node} $c$ represents a product $\prod_{c \in \text{in}(n)} h_c(\X_c)$. 
The scope of a product node is the union of the scope of its input nodes, $\bigcup_{c' \in \text{in}(c)} \X_{c'}$.
For tractable inference, it is required that a circuit is \emph{structured decomposable} (\cref{def:structured_decomposability}). 
A circuit is decomposable if all inputs of product units have \emph{disjoint scopes}, so that $\X_{c'} \cap \X_{c''} = \emptyset \ \forall c', c'' \in \text{in}(c)$. 
For structural decomposability, it is further required that all pairs of product nodes $n, m$ with same scope decompose the scope in the same way, so that $\X_{\text{in}_i(n)} = \X_{\text{in}_i(m)} \ \forall i \in \{1, \dots, C \}$, where $C$ is the number of inputs in $\text{in}(n)$ and $\text{in}(m)$.

\paragraph{Sum node} 

A \emph{sum node} computes the sum $s(\X_c) = \sum_{c' \in \text{in}(c)} w_{c, c'} h_{c'}(\X_{c'})$, where $\w_n = [w_{c, 1}, \dots, w_{c, C}]$ are the trainable parameters of the node and $C$ are the number of inputs of node $c$. 
All sum nodes in a circuit have to satisfy the structural condition \emph{smoothness} for tractable inference. 
A circuit is smooth if, for all sum nodes, the inputs of a sum node have \emph{same scope}, so $\X_c = \X_{c'} \ \forall c, c' \in \text{in}(c)$.

If a circuit is \emph{structured decomposable}, statistical moments of $f(\X)$ can be computed by decomposing the complex integral into smaller, easier to solve integrals based on the observation that integration interchanges with summation and integration interchanges with decomposable multiplication \citep{choi2020Probabilistic}.
Essentially, the integral gets ``pushed down'' to the input nodes of the circuit, which can be evaluated easily. 
It is then sufficient to perform a forward pass through the circuit using the evaluated integrals at the circuit leaves to compute the desired statistical moment of the whole circuit $f(\X)$.

In tensorized circuit representations, multiple nodes are parallelized and grouped to circuit \emph{layers}.
Usually, a layer includes multiple nodes encoding the \emph{same function} over the \emph{same scope}, but each with a distinct set of parameters \citep{peharz2020Einsum}, increasing the number of parameters and expressivity of the circuit. 
The layerwise representation yields two types of product layers: (i) product layers that compute the \emph{outer product} $\mathbf{A}_{N \times N} = \mathbf{u} \otimes \mathbf{v}$, where $\mathbf{u}$ is the vector of all $N$ outputs from children nodes with scope $\X_c$ and $\mathbf{v}$ is the vector of all $N$ outputs from children nodes with scope $\X_{c'}$, also referred to in the circuit literature as a \emph{Kronecker product} \citep{loconte2025What}, (ii) product layers computing the \emph{Hadamard product} $\mathbf{a} = \mathbf{u} \odot \mathbf{v}$.
Aside from the PCE input layer computing the tensor product $\TP$, which requires a Kronecker product layer to compute all expansion terms of the full tensor product basis, both Hadamard or Kronecker product layers can be used.

\section{TRACTABLE INFERENCE}
\label{app:inference}

In this section, we derive the tractable computations of statistical moments in DeepPCE via simple forward passes, leveraging PCE and circuit properties.

\subsection{Expectations}
\label{app:inference_expectation}

First, we want to compute the expectation of the whole circuit $f(\X) = \Y$:

\begin{equation}
    \E[\Y] = \int p(\x) f(\x) \text{d}\x
\end{equation}

We first compute the expectation at the input nodes of the circuit and show how we can pass them through product and sum nodes to get the result at the output.

\subsubsection*{PCE input layer}
\label{app:inference_expectation_input}

We recall that, following from the orthogonality property of PCE (\cref{eq:orthonormal_tensor_products}), the expectation of a single input node PCE $g_n$ over scope $\X_c$ is

\begin{equation}
\label{eq:app_pce_expectation}
    \E[g_n(\X_c)] = w_{n, \midxa_0}
\end{equation}

To compute the expectation for an input node, we thus simply have to set all $w_{n, \midxa_i} = 0 \ \forall \ i > 0$.

\subsubsection*{Product layer}
\label{app:inference_expectation_product}

The DeepPCE is a \emph{decomposable} circuit --- inputs to product nodes always have disjoint scopes $\X_c \cap \X_{c'} = \emptyset$. 
The product $p_{n, m}$ of two inputs $g_n, g_m$ is

\begin{equation}
    p_{n, m} = g_n(\X_c) g_m(\X_{c'})
\end{equation}

Because the circuit is decomposable and the input distribution $p(\X)$ factorizes, we get that $g_n(\X_c) \ {\perp \!\!\! \perp} \ g_m(\X_{c'})$. 
We can thus decompose the expectation of the product:

\begin{align}
\label{eq:app_product_layer_expectation}
    \E[p_{n, m}] &= \E[g_n(\X_c) g_m(\X_{c'})] \nonumber \\
          &= \E[g_n(\X_c)] \E[g_m(\X_{c'})] \qquad g_n(\X_c) \ {\perp \!\!\! \perp} \ g_m(\X_{c'})
\end{align}

We only have to pass the expectations computed in \cref{app:inference_expectation_input} to the product node to compute its expectation.

\subsubsection*{Sum layer}

In case of an outer product layer (sec. \cref{app:circuit_preliminaries}), a sum node with index $k$ computes the sum 

\begin{equation}
    s_{k} = \sum_{n, m} w_{k, n, m} p_{n, m}
\end{equation}

The expectation is then

\begin{equation}
\label{eq:app_sum_layer_expectation}
    \E[s_{k}] = \sum_{n, m} w_{k, n, m} \E[p_{n, m}]
\end{equation}

We know the expectation of the product $\E[p_{n, m}]$ from \cref{app:inference_expectation_product}. 
As deeper circuits are only sequences of sum and product nodes, it sufficies to compute the expectations at the input nodes and perform a forward pass through the circuit to compute $\E[\Y]$.

\subsection{Covariances}
\label{app:inference_covariance}

Next we want to compute the covariance of all DeepPCE outputs $\cov(\Y, \Y)$, which is a $O \times O$ positive definite matrix. 
We show how covariances are computed at the input layer can be propagated through the network.

\subsubsection*{PCE input layer}

First, we compute the covariances at the input nodes, which are:

\begin{equation}
\label{eq:app_covariance}
    \cov(g_n(\X_c), g_m(\X_{c'})) = \E[g_n(\X_c) \  g_m(\X_{c'})] - \E[g_n(\X_{c'})] \E[g_m(\X_{c'})]
\end{equation}

Recall that we assume a factorized distribution of the inputs $\X$. 
The covariance for all pairs of input node PCEs $g_n(\X_c), g_m(\X_{c'})$ with disjoint scopes $\X_c \neq \X_{c'}$ is thus zero, so we only have to compute the covariances from input nodes that share the same scope. 
We know how to compute $\E[g_n(\X_c)]$ and $\E[g_m(\X_c)]$ from \cref{eq:app_pce_expectation}. 
For a PCE input node, we can compute the second moment, i.e. the expectation of a product of two PCEs $g_n, g_m$, over the same scope as

\begin{align}
\E[g_n(\X_c) \  g_m(\X_c)] &= \E \left[  \sum_{\midxa \in \mathcal{A}_c} \sum_{\midxb \in \mathcal{A}_c} w_{n, \midxa} w_{m, \midxb} \TP_{\midxa}(\X_c) \TP_{\midxb}(\X_c) \right] \nonumber \\
&= \sum_{\midxa, \midxb \in \mathcal{A}_c} w_{n, \midxa} w_{m, \midxb} \underbrace{\E[\TP_{\midxa}(\X_{c, \mathcal{I}}) \TP_{\midxb}(\X_{c, \mathcal{I}})]}_{= \delta_{\midxa, \midxb}}
\end{align}

where $\midxa$ is the multi-index of the PCE $g_n$ and $\midxb$ is the multi-index of the PCE $g_m$. 
We know again from the orthogonality property that $\E[\TP_{\midxa}(\X) \TP_{\midxb}(\X)] = \delta_{\midxa, \midxb}$, where $\delta_{\midxa, \midxb}$ is the Kronecker delta. 
The second moment is thus

\begin{equation}
\label{eq:app_pce_second_moment}
\E[g_n(\X_c) \ g_m(\X_c)] = \sum_{\midxa \in \mathcal{A}_c} w_{n, \midxa} w_{m, \midxa}
\end{equation}

We get the covariance using the result for the expectation from \cref{eq:app_pce_expectation}:

\begin{align}
\label{eq:app_pce_covariance}
\cov(g_n(\X_c), g_m(\X_c)) &= \E[g_n(\X_c) g_m(\X_c)] - \E[g_n(\X_c)] \E[g_m(\X_c)] \nonumber \\
&= \sum_{\midxa \in \mathcal{A}_c} w_{n, \midxa} w_{m, \midxa} - w_{n, \midxa_0} w_{m, \midxa_0}\nonumber \\ 
&=\sum_{\midxa \in \mathcal{A}_c \setminus \{\midxa_0\}} \ \sum_{\midxb \in \mathcal{A}_c \setminus \{\midxb_0\}} w_{n, \midxa} w_{m, \midxb}
\end{align}

\subsubsection*{Product layer}

To compute second moments at the product layer, we again apply $g(\X_c) \ {\perp \!\!\! \perp} \ g(\X_{c'})$ due to decomposability:

\begin{align}
\label{eq:app_product_layer_second_moment}
    \E[p_{n, m} p_{k, l}] &= \E [g_n(\X_c) g_m(\X_{c'}) g_k(\X_c) g_l(\X_{c'})] \nonumber \\
    &= \E [g_n(\X_c) g_k(\X_c)] \E[ g_m(\X_{c'}) g_l(\X_{c'})]
\end{align}

We know how to compute the expectation of products over the same scope from \cref{eq:app_pce_second_moment}. 
We can now compute the covariance

\begin{align}
    \cov(p_{n, m}, p_{k, l}) &= \E[p_{n, m} p_{k, l}] - \E[p_{n, m}] \E[p_{k, l}]
\end{align}

by applying \cref{eq:app_product_layer_second_moment} and \cref{eq:app_product_layer_expectation}. 
We are thus able to compute second moments at a product node by propagating second moments from the input nodes.
Covariances can be computed similarly, with the additional need of computing expectations.

\subsubsection*{Sum layer}

Second moments of two sums $s_i, s_j$ over same scope are

\begin{align}
\label{eq:app_sum_layer_second_moment}
    \E[s_{i} s_{j}] &= \E \left[ \sum_{n, m} w_{i, n, m} p_{n, m} \sum_{k, l} w_{j, k, l} p_{k, l} \right] \nonumber \\
    &= \sum_{n, m, k, l} w_{i, n, m} w_{j, k, l} \E \left[ p_{n, m} p_{k, l} \right] 
\end{align}

We know how to compute the expectation of the product of two product nodes from \cref{eq:app_product_layer_second_moment}. 
We finally are able to compute the covariance

\begin{align}
    \cov(s_i, s_j) =  \E[s_{i} s_{j}] - \E[s_i]  \E[s_{j}]
\end{align}

using the result for the second moment of sums in \cref{eq:app_sum_layer_second_moment} and the expectation of sums in \cref{eq:app_sum_layer_expectation}. 
We show that statistical queries for the second moment $\E[\Y \Y]$ and the covariance $\cov(\Y, \Y)$ can be answered by computing those quantities at the input node PCEs. 
Leveraging the orthogonality property, they reduce to simple operations only involving the learned weights of input nodes.

\subsection{Conditional expectations}
\label{app:inference_conditional_expectation}

Now, we consider the case of the conditional expectation $\E[\Y | \x_{\mathcal{I}}]$, where $\mathcal{I} \subseteq \{{1, \dots, D}\}$ is an index set of some parameters in $\X$. 
The set of marginal variables is denoted as $\X_{\neg \mathcal{I}}$ with $\neg \mathcal{I} = \{1, \dots, D\} \setminus \mathcal{I}$, so that $\X_{\mathcal{I}} \cup \X_{\neg \mathcal{I}} = \X$. 
The conditional expectation is

\begin{align}
\E_{\X_{\neg \mathcal{I}}}[{\Y | \x_{\mathcal{I}}}] &= \int p(\x_{\neg \mathcal{I}} | \x_{\mathcal{I}}) f(\overbrace{\x_{\neg \mathcal{I}}, \x_{\mathcal{I}}}^{\x}) \text{d}\x_{\neg \mathcal{I}} \nonumber \\
&= \int p(\x_{\neg \mathcal{I}}) f(\x_{\neg \mathcal{I}}, \x_{\mathcal{I}}) \text{d}\x_{\neg \mathcal{I}}
\end{align}

The second step again follows from the assumption that $p(\X)$ factorizes, so that $p(\x_{\neg \mathcal{I}} | \x_{\mathcal{I}}) = p(\x_{\neg \mathcal{I}})$.
We can write the conditional expectation of a PCE at the input layer as

\begin{equation}
\label{eq:app_conditional_expectation_fixed}
\E_{\X_{c, \neg \mathcal{I}}}[{g_n(\X_c) | \x_{c, \mathcal{I}}}] = \E_{\X_{c, \neg \mathcal{I}}}\left[\sum_{\midxa \in \mathcal{A}_c} w_{n, \midxa} \TP_{\midxa}(\X_{c, \neg \mathcal{I}}) \hat{\TP}_{\midxa}(\x_{c, \mathcal{I}})\right]
\end{equation}

where $\hat{\TP}_{\midxa}(\x_{c, \neg \mathcal{I}})$ is the tensor product of the conditional variables:

\begin{equation}
\hat{\TP}_{\midxa}(\x_{c, \mathcal{I}}) = \prod_{x_{e} \in \x_{c, \mathcal{I}}} \psi_{\alpha_e}(x_{e})
\end{equation}

Based on \cref{eq:app_conditional_expectation_fixed}, we get

\begin{align}
\label{eq:app_conditional_expectation_fixed2}
\E_{\X_{c, \neg \mathcal{I}}}[{g_n(\X_c) | \x_{c, \mathcal{I}}}] &= \sum_{\midxa \in \mathcal{A}_c} w_{n, \midxa} \hat{\TP}_{\midxa}(\x_{c, \mathcal{I}}) \underbrace{\E[\TP_{\midxa}(\X_{c, \neg \mathcal{I}})]}_{\substack{= 1 \text{ if } \alpha_{i} = 0 \; \forall \; \alpha_{i} \, \in \, \midxa, \\ \text{ else } 0}} \nonumber \\
&= \sum_{\midxa \in \mathcal{A}_{c, \mathcal{I}}} w_{n, \midxa} \hat{\TP}_{\midxa}(\x_{c, \mathcal{I}}) \quad \text{with } \mathcal{A}_{c, \mathcal{I}} = \{ \midxa: \ \forall j \in \neg \mathcal{I}: \alpha_j = 0 \}
\end{align}

It follows from the orthogonality property of the PCE that $\E[\TP_{\midxa}(\X_{c, \neg \mathcal{I}})] = 1$ only if $\{ \midxa: \ \forall j \in \neg \mathcal{I}: \alpha_j = 0 \}$. 
To compute the conditional expectation, we set all input layer PCE weights to zero not meeting this requirement.
Recycling our computations for expectations in product and sum layers (\cref{app:inference_expectation}), we only have to perform a forward pass to compute $\E_{\X_{c, \neg \mathcal{I}}}[\Y | \x_{\mathcal{I}}]$.

\subsection{Conditional covariances}
\label{app:inference_conditional_covariance}

We compute the conditial covariance $\cov_{\X_{\neg \mathcal{I}}}(\Y, \Y | \x_{\mathcal{I}})$ using the same method as described in \cref{app:inference_covariance}. 
The PCE conditional second moment is

\begin{align}
\E_{\X_{c, \neg \mathcal{I}}}[g_n(\X_{c, \neg \mathcal{I}} | \x_{\mathcal{I}}) &\  g_m(\X_{c, \neg \mathcal{I}} | \x_{\mathcal{I}})] = \nonumber \\ 
&= \E_{\X_{c, \neg \mathcal{I}}} \left[  \sum_{\midxa, \midxb \in \mathcal{A}_c} w_{n, \midxa} w_{m, \midxb} \hat{\TP}_{\midxa}(\x_{\mathcal{I}}) \hat{\TP}_{\midxb}(\x_{\mathcal{I}}) \TP_{\midxa}(\X_{c, \neg \mathcal{I}}) \TP_{\midxb}(\X_{c, \neg \mathcal{I}}) \right] \nonumber \\
&= \sum_{\midxa, \midxb \in \mathcal{A}_c} w_{n, \midxa} w_{m, \midxb} \hat{\TP}_{\midxa}(\x_{\mathcal{I}}) \hat{\TP}_{\midxb}(\x_{\mathcal{I}}) \underbrace{\E \left[\TP_{\midxa}(\X_{c, \neg \mathcal{I}}) \TP_{\midxb}(\X_{c, \neg \mathcal{I}})\right]}_{= \delta_{\midxa, \midxb}} \nonumber \\
&= \sum_{\midxa \in \mathcal{A}_c \setminus \{\midxa_0\}} \ \sum_{\midxb \in \mathcal{A}_c \setminus \{\midxb_0\}} w_{n, \midxa} w_{m, \midxb} \hat{\TP}_{\midxa}(\x_{\mathcal{I}}) \hat{\TP}_{\midxb}(\x_{\mathcal{I}})
\end{align}

We can now compute the covariance using \cref{eq:app_covariance}. 
Based on \cref{app:inference_covariance}, we perform a forward pass to compute $\cov_{\X_{\neg \mathcal{I}}}(\Y, \Y | \x_{\mathcal{I}})$.

\subsection{Expectations of covariances and covariances of expectations}

Additional to computing statistical moments, we are interested in performing sensitivity analysis by identifying the input variables $X_d \in \X$ that contribute most to the variance of $f(\X) = \Y$ represented by the DeepPCE.

Recalling the definition of Sobol indices from \cref{eq:def-first-order-sobol,eq:def-total-order-sobol}, we are ultimately interested in computing the covariance of expectations conditioned on the random variables $\X_{\mathcal{I}} \subset \X$,  $\cov_{\X_{\mathcal{I}}}(\E_{\X_{\neg \mathcal{I}}}[{\Y | \X_{\mathcal{I}}}])$ as well as the expectation of conditional covariances: $\E_{\X_{\mathcal{I}}}[\cov_{\X_{\neg \mathcal{I}}}({\Y | \X_{\mathcal{I}}})]$, quantifying the expectation or covariance based \emph{only} on the randomness of $\X_{\mathcal{I}}$.
We again apply the standard formula for the covariance

\begin{align}\label{eq:app_covariance_of_conditional_expectations}
    \cov_{\X_{\mathcal{I}}}(\E_{\X_{\neg \mathcal{I}}}[{\Y | \X_{\mathcal{I}}}], \E_{\X_{\neg \mathcal{I}}}[{\Y | \X_{\mathcal{I}}}]) &= \E_{\X_{\mathcal{I}}}[\E_{\X_{\neg \mathcal{I}}}[{\Y | \X_{\mathcal{I}}}] \ \E_{\X_{\neg \mathcal{I}}}[{\Y | \X_{\mathcal{I}}}]] - \E_{\X_{\mathcal{I}}}[\E_{\X_{\neg \mathcal{I}}}[{\Y | \X_{\mathcal{I}}}]]^2 \nonumber \\
    &= \E_{\X_{\mathcal{I}}}[\E_{\X_{\neg \mathcal{I}}}[{\Y | \X_{\mathcal{I}}}] \ \E_{\X_{\neg \mathcal{I}}}[{\Y | \X_{\mathcal{I}}}]] - \E[\Y]^2
\end{align}

For the simplification in the second step, we use the law of total expectation, which states that $\E[\E[Y | X]] = \E[Y]$. 
We know how to compute $\E[\Y]$ based on the derivation in \cref{app:inference_expectation}. 
We further apply our results from \cref{eq:app_conditional_expectation_fixed2} to compute the second moment of the distribution of the conditional expectation at the PCE nodes:

\begin{align}
    \E_{\X_{c, \mathcal{I}}} \Big[ \E_{\X_{c, \neg \mathcal{I}}}[{g_n(\X_{c, \neg \mathcal{I}}) | \X_{c, \mathcal{I}}}] &\ \E_{\X_{c, \neg \mathcal{I}}}[{g_m(\X_{c, \neg \mathcal{I}}) | \X_{c, \mathcal{I}}}] \Big] = \nonumber \\ 
    &= \E \left[ \sum_{\midxa \in \mathcal{A}_{c, \mathcal{I}}} w_{n, \midxa} \hat{\TP}_{\midxa}(\X_{\mathcal{I}}) \sum_{\midxb \in \mathcal{A}_{c, \mathcal{I}}} w_{m, \midxb} \hat{\TP}_{\midxb}(\X_{\mathcal{I}}) \right] \nonumber \\
    &= \sum_{\midxa, \midxb \in \mathcal{A}_{c, \mathcal{I}}} w_{n, \midxa} w_{m, \midxb} \underbrace{\E \left[ \hat{\TP}_{\midxa}(\X_{\mathcal{I}}) \hat{\TP}_{\midxb}(\X_{\mathcal{I}}) \right]}_{= \delta_{\midxa, \midxb}}
\end{align}

where $\mathcal{A}_{c, \mathcal{I}} = \{ \midxa: \ \forall j \in \neg \mathcal{I}: \alpha_j = 0 \}$ as defined in \cref{eq:app_conditional_expectation_fixed2}. 
Based the orthogonality property of the PCE, the equation reduces to

\begin{align}
\label{eq:app_second_moment_conditional_expectation}
    \E_{\X_{c, \mathcal{I}}} \Big[ \E_{\X_{c, \neg \mathcal{I}}}[{g_n(\X_{c, \neg \mathcal{I}} | \X_{c, \mathcal{I}})}] &\ \E_{\X_{c, \neg \mathcal{I}}}[{g_m(\X_{c, \neg \mathcal{I}} | \X_{c, \mathcal{I}})}] \Big] = \nonumber \\ 
    &= \sum_{\midxa \in \mathcal{A}_{c, \mathcal{I}}} w_{n, \midxa} w_{m, \midxa} \underbrace{\E \left[ \hat{\TP}_{\midxa}(\X_{\mathcal{I}}) \hat{\TP}_{\midxa}(\X_{\mathcal{I}}) \right]}_{= 1} \nonumber \\
    &= \sum_{\midxa \in \mathcal{A}_{c, \mathcal{I}}} w_{n, \midxa} w_{m, \midxa}
\end{align}

We are now able to compute the covariance of conditional expectations at the PCE nodes according to \cref{eq:app_covariance_of_conditional_expectations} by inserting \cref{eq:app_second_moment_conditional_expectation}:

\begin{align}
\label{eq:app_covariance_conditional_expectation}
    \cov_{\X_{c, \mathcal{I}}}(\E_{\X_{c, \neg \mathcal{I}}}[{g_n(\X_c) | \X_{c, \mathcal{I}}}], \E_{\X_{c, \neg \mathcal{I}}}[{g_n(\X_c) | \X_{c, \mathcal{I}}}]) &= \sum_{\midxa \in \mathcal{A}_{c, \mathcal{I}}} w_{n, \midxa} w_{m, \midxa} - w_{n, \midxa_0} w_{m, \midxa_0} \nonumber \\
    &= \sum_{\midxa \in \mathcal{A}_{c, \mathcal{I}} \setminus \{\midxa_0\}} w_{n, \midxa} w_{m, \midxa}
\end{align}

Finally, we are also able to compute the expectation of conditional covariances using the law of total covariance, stating that $\cov(X, Y) = \E[\cov(X, Y | Z)] + \cov(\E[X | Z], \E[Y | Z])$:

\begin{equation}
\label{eq:app_law_of_total_covariance}
   \E_{\X_{\mathcal{I}}} [\cov_{\X_{\neg \mathcal{I}}}(\Y, \Y | \X_{\mathcal{I}})] = \cov(\Y, \Y) - \cov_{\X_{\mathcal{I}}}(\E_{\X_{\neg \mathcal{I}}}[{\Y | \X_{\mathcal{I}}}], \E_{\X_{\neg \mathcal{I}}}[{\Y | \X_{\mathcal{I}}}]),
\end{equation}

by computing the covariance according to \cref{eq:app_covariance} and the covariance of conditional expectations according to \cref{eq:app_covariance_conditional_expectation}.

\section{VALIDATION OF MODEL INFERENCE}

\begin{figure}
    \centering
    \includegraphics[width=\textwidth]{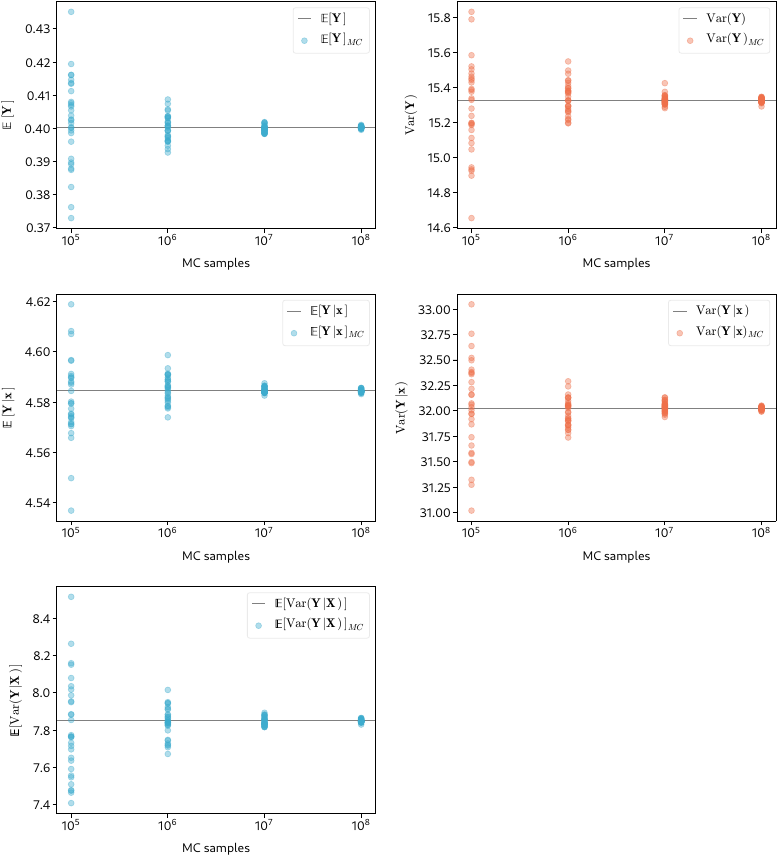}
    \caption{Inference pass evaluation comparing the analytical solution (grey line) to Monte Carlo approximations (blue/orange circles). The first row show unconditional expectation $\E[Y]$ (left) and unconditional variance $\Var(Y)$, the second row conditional expectation $\E[Y | \x]$ and conditional variance $\Var(Y | \x)$, the third row shows the expectation of the conditional variance $\E[\Var(Y | \X_{\mathcal{I}})]$}
    \label{fig:mc_tests}
\end{figure}

We test all inference queries by comparing the analytical results with Monte Carlo approximations of those quantities from the same DeepPCE. 
We compute Monte Carlo approximations with different sample sizes $S = [10^5, 10^6, 10^7, 10^8]$, performing 30 individual Monte Carlo runs per sample size setting. 
We then test the convergence with increasing sample size and compare it to the analytical solution derived above by performing one sample t-tests for all outputs, showing that the difference between the analytical solutions and the Monte Carlo approximations is not statistical significant for any outputs (\cref{tab:p_values_mc}). 
\cref{fig:mc_tests} shows results for a DeepPCE with 8 inputs $\X \sim \mathcal{N}(\mathbf{0}, \mathbf{I})$ and 8 outputs, from which we randomly chose a single output for visualization. 
We iteratively validate the analytical solutions and use them in further Monte Carlo tests as described below.

\subsubsection*{Expectation and covariance}

We first test the result of the analytical computation obtained with the model, $\E[\Y] = \E[\mathcal{M}(\X)] $, by comparing it with the Monte Carlo estimate $\E[\Y]_{MC}$:

\begin{equation}
\label{eq:app_mc_expectation}
    \E[\Y]_{MC} = \frac{1}{n} \sum_i^N \mathcal{M}(\X)
\end{equation}

After confirming that $\E[\Y] = \E[\Y]_{MC}$, we apply the analytical expectation in the Monte Carlo approximation of the covariance:

\begin{equation}
    \text{cov}(\Y, \Y)_{MC} = \frac{1}{n - 1} \sum_i^N (\mathcal{M}(\X) - \underbrace{\E[\Y]}_{\text{tested}})^2
\end{equation}

\subsubsection*{Conditional expectation and conditional covariance}

For the validation of conditional moments, we condition on variables $\X_{\mathcal{I}} = \{X_1, X_2, X_3, X_5 \}$. 
For tests with fixed conditional values, we chose conditioned values as $\X_{\mathcal{I}} \sim \mathcal{U}(-2, 2)$. 
We test the conditional expectation similar to \cref{eq:app_mc_expectation}:

\begin{equation}
    \E[\Y | \x_{\mathcal{I}}]_{MC} = \frac{1}{n} \sum_i^N \mathcal{M}(\X_{\neg {\mathcal{I}}}, \x_{\mathcal{I}})
\end{equation}

After confirming the results, we use it for the Monte Carlo approximation of the conditional covariance

\begin{equation}
    \text{cov}(\Y, \Y | \x_{\mathcal{I}})_{MC} = \frac{1}{n - 1} \sum_i^N (\mathcal{M}(\X_{\neg \mathcal{I}}, \x_{\mathcal{I}}) - \underbrace{\E[\Y | \x_{\mathcal{I}}])^2}_{\text{tested}}
\end{equation}

The Monte Carlo approximation for the expected conditional covariance is then simply

\begin{equation}
    \E[\text{cov}(\Y, \Y | \X_{\mathcal{I}})]_{MC} = \frac{1}{n} \sum_i^N \underbrace{\text{cov}(\Y, \Y | \x_i)}_{\text{tested}}
\end{equation}

Because we get the covariance of conditional expectations just from known quantities (\cref{eq:app_law_of_total_covariance}), we do not have to perform a separate evaluation with Monte Carlo.

\begin{table}[b]
    \caption{P-values computed using one sample t-tests, comparing the analytical solution of the statistical inferences to Monte Carlo approximations. The results are not statistical significant for any outputs, indicating that the analytical solution is not different from the Monte Carlo solution.\\}
    \centering
    \begin{tabular}{cccccc}
\toprule
Outputs & \multicolumn{5}{c}{p-values} \\
\cmidrule(lr){1-1} \cmidrule(lr){2-6}
 & $\mathbb{E} \; [Y]$ & $\mathrm{Var}(Y)$ & $\mathbb{E} \; [Y | x]$ & $\mathrm{Var}(Y | x)$ & $\mathbb{E} \; [\mathrm{Var}(Y | X)]$ \\
 \cmidrule(lr){2-2} \cmidrule(lr){3-3} \cmidrule(lr){4-4} \cmidrule(lr){5-5} \cmidrule(lr){6-6}
$Y_1$ & 0.875 & 0.950 & 0.581 & 0.172 & 0.493 \\
$Y_2$ & 0.300 & 0.119 & 0.575 & 0.583 & 0.986 \\
$Y_3$ & 0.108 & 0.158 & 0.643 & 0.660 & 0.873 \\
$Y_4$ & 0.132 & 0.223 & 0.706 & 0.495 & 0.970 \\
$Y_5$ & 0.915 & 0.816 & 0.521 & 0.204 & 0.522 \\
$Y_6$ & 0.093 & 0.058 & 0.679 & 0.776 & 0.916 \\
$Y_7$ & 0.382 & 0.148 & 0.567 & 0.559 & 0.951 \\
$Y_8$ & 0.250 & 0.104 & 0.591 & 0.610 & 0.964 \\
\bottomrule
\end{tabular}

    \label{tab:p_values_mc}
\end{table}

\newpage
\section{BATCH NORMALIZATION AND ORTHOGONALITY}
\label{app:batch_norm}

We show that orthogonality is preserved in DeepPCE also when batch normalization is applied at the sum layers. 
Batch normalization, denoted as the function $z$, applies an affine transformation at inference time:

\begin{equation}
        z(\Y) = \gamma \frac{\Y - \E[\Y]}{\sqrt{\text{Var}(\Y) + \varepsilon}} + \beta
\end{equation}

where $\Y$ is previous hidden layer's output, $\gamma$ and $\beta$ are learnable parameters of the batch norm layer and $\E[\Y]$ and $\Var(\Y)$ are the estimates for mean and variance respectively and $\varepsilon$ is a small constant added for numerical stability. 
At inference time, the batch statistics are independent from the inputs. 
The equation can be reformulated as

\begin{gather}
     z(\Y)= \hat{\gamma} \Y + \hat{\beta} \\
     \text{with } \hat{\gamma} = \frac{\gamma}{\sqrt{\Var[\Y] + \varepsilon}}, \quad
    \hat{\beta} = \beta - \frac{\gamma \E[\Y]}{\sqrt{\Var[\Y] + \varepsilon}}
\end{gather}

Applied to a PCE $f(\X) = \sum_{\midxa \in \mathcal{A}} w_{\midxa} \TP_{\midxa} (\X)$ we get:

\begin{equation}
  z(f(\X)) =  \hat{\gamma} \bigg(\sum_{\midxa \in \mathcal{A}} w_{\midxa} \TP_{\midxa}(\X)\bigg) + \hat{\beta}
\end{equation}

The batch norm coefficients $\gamma$ and $\beta$ can be subsumed in the weights and the zeroth-order term of the expansion, i.e. the constant $w_{\midxa_0}$, respectively, thus not affecting orthogonality:

\begin{equation}
  z(f(\X)) = \sum_{\midxa \in \mathcal{A}} \hat{\gamma} \ w_{\midxa} \TP_{\midxa}(\X) + b_{\midxa}
 \quad \text{with }
    b_{\midxa} = \begin{cases}
        \hat{\beta}    & \text{if } \alpha_i = 0 \ \forall \alpha_i \in \midxa \\
        0        & \text{otherwise}
            \end{cases} \\
\end{equation}

\section{EXPERIMENTS}
\label{app:experiments}

\subsection{Benchmark functions}

We adopt benchmark functions that are widely used in the uncertainty quantification and sensitivity analysis literature. 
The Sobol G* and Bratley Sum functions are our own re-implementations of examples from the comprehensive collection of test functions in the \texttt{uqtestfuns} library \citep{wicaksono2023UQTestFuns}---we refer the reader there for more detailed information.

\subsubsection{Sobol G*}

The Sobol G* function \citep{saltelli1995Use} is defined as

\begin{gather}
    \label{eq:sobolgstar}
    f(\X, \mathbf{a}, \boldsymbol{\delta}, \boldsymbol{\alpha}) = \prod_{i=1}^D g_k (x_i, a_i, \delta_i, \alpha_i) \\
    \text{where} \quad  g_k (x_i, a_i, \delta_i, \alpha_i) = \frac{(1 + \alpha_i) | 2 (x_i + \delta_i - \lfloor x_i + \delta_i \rfloor) - 1 |^{\alpha_i} + a_i}{1 + a_i}, \nonumber
\end{gather}

with $\X \sim \mathcal{U}[0, 1]^d$. Each parameter $a_i \in \mathbf{a} \geq 0$ controls the ``importance'' of the input variable $X_i \in \X$. 
Smaller values of $a_i$ translate to a higher contribution of $X_i$ to the global variance $\Var(\X)$. 
We generate $\mathbf{a}$ with $a_i = \frac{i - 1}{D - 1} \cdot 10$. 
$\boldsymbol{\delta} \sim \mathcal{U}[0, 1]^d$ is a shift parameter that influences the shape of the function but not the sensitivity indices and is thus sampled randomly. 
$\boldsymbol{\alpha} = \mathbf{\frac{1}{2}}$ controls the curvature of the function. 
The floor operator introduces additional nonlinearity and makes this function an interesting test case. 
Analytical solutions to first order and total effect Sobol indices are known.

\subsubsection{Bratley Sum}

This function was introduced in \citet{bratley1992Implementation} and later used in \citet{kucherenko2009Monte} and \citet{saltelli2010Variance}:

\begin{equation}
    f(\X) = \sum_{i = 1}^D (-1)^i \prod_{j = 1}^i x_i
\end{equation}

The joint input distribution is a uniform distribution on the unit hypercube, $\X \sim \mathcal{U}[0, 1]^d$. 
This function features only low-order interactions and favors approximations using sparse, strongly truncated bases. 
Analytical solution for the total effect Sobol indices are known from \citet{saltelli2010Variance}.

\subsubsection{PCE benchmark}

This function from \citet{luthen2021Sparse} is usually used with $D = 100$:

\begin{equation}
\label{eq:pce_benchmark}
    f(\X) = 3 - \frac{5}{d} \sum_{i = 1}^d i X_i + \frac{1}{d} \sum_{i = 1}^d i X^3_i + \frac{1}{3 d} \sum_{i = 1}^d i \ln(X^2_i + X^4_i) + X_1 X_2^2 + X_2 X_4 - X_3 X_5 + X_{51} + X_{50} X^2_{54}  
\end{equation}

where inputs $\X = \{ X_1, \dots X_D \}$ are sampled from uniform distributions $X_i \sim \mathcal{U}(1, 2) \ \forall i \neq 20$ and $X_{20} \sim \mathcal{U}(1, 3)$. 
This test function is interesting due to different intervals of the input distributions and the presence of interaction variables. 
It is usually used to benchmark sparse PCEs, as only low-order interactions are present.

\subsection{Experiment details}

Each experiment was performed in isolation on a SLURM cluster, using a single NVIDIA A40 GPU (48 GB Memory) with 2 CPU cores and 64 GB memory.
We peformed each experiment 10 times for each model to provide estimates of error variances.
The DeepPCE, MLP and UNet were trained for a maximum of 500 epochs, using early stopping to prevent overfitting.
As described in \cref{sec:training}, the DeepPCE is susceptible to bad parameter initializations. 
To mitigate this, we initialized 10 DeepPCEs and continued training with the best one after 50 epochs.
This resulted in no runs converging to bad local minima.
Due to the consistent performance of MLPs and UNets, this training procedure was not necessary for those two model classes.
For all experiments, we report relative MSEs as

\begin{equation}
\label{eq:app_rel_mse}
    R = \frac{\frac{1}{N}\sum_i^N(y  - \hat{y})^2}{\frac{1}{N}\sum_i^Ny^2}
\end{equation}

where $y$ are the true values and $\hat{y}$ are the model predictions.

\subsection{Hyperparameters}

For each experiment, the DeepPCE, MLP and UNet were tuned using the same computational budget.
We tuned the learning rate (lr) for all models, using values $[0.001, 0.005, 0.01, 0.05, 0.1]$ and a fixed batch size of 128.
Additional tuned parameters for the DeepPCE include the number of sum nodes per layer (num sums), the maximum order of the polynomial expansion (max order), the standard deviation of the Gaussian distribution used for weight initialization (std) as well as the variance decay parameter ($\lambda$) described in \cref{eq:weight_init_scaling}.
We used a fixed input node scope size of 1 and Hadamard product layers.
Additional MLP tuning parameters include the number of hidden layers (num hidden) and the number of units per hidden layer (num units), specific UNet tuning parameters are the number of channels at the bottleneck of the encoder (num channels). 
The hyperparameters for all experiments are shown in \cref{tab:hyperparams}, \cref{tab:model_params} shows details on model sizes and training time. 

For the sparse PCE variants ($\text{PCE}_{OMP}$, $\text{BA-PCE}_{OMP}$), the stopping criterion for the orthogonal matching pursuit algorithm is the maximum number of active dimensions $A$, with $A \approx \frac{1}{10} D$, where $D$ is the number of input dimensions.

\begin{table}
   \caption{Experiment hyperparameter settings.}
    \centering
    \begin{tabular}{l  lr  lr  lr}
   & \multicolumn{2}{c}{DeepPCE} & \multicolumn{2}{c}{MLP} & \multicolumn{2}{c}{UNet}\\
   \midrule
   \multirow{5}{*}{Darcyflow} & lr & 0.005   &   lr & 0.001 &  lr  &  0.005 \\
   & num sums & 100  & num hidden & 2                       & num channels & 1024\\     
   & max order & 3     & num units & 2048  & & \\
   & std & 0.9     &   &  & & \\
   & $\lambda$ & 0.3     &   &   & & \\

   \cmidrule{1-7}
   \multirow{5}{*}{Steady state diffusion} & lr & 0.001   &   lr & 0.001 &  lr  &  0.001 \\
   & num sums & 50  & num hidden & 2                       & num channels & 1024\\     
   & max order & 3     & num units & 2048  & & \\
   & std & 0.15     &   &  & & \\
   & $\lambda$ & 0.15     &   &   & & \\

   \cmidrule{1-5}
   \multirow{5}{*}{Sobol G*} & lr & 0.01  &   lr & 0.001 &   &  \\
   & num sums & 20  & num hidden & 3                       &   & \\     
   & max order & 6     & num units & 512  & & \\
   & std & 0.7     &   &  & & \\
   & $\lambda$ & 0.15    &   &   & & \\

   \cmidrule{1-5}
   \multirow{5}{*}{XD Benchmark} & lr & 0.005   &   lr & 0.001 &    &  \\
   & num sums & 20  & num hidden & 2                       &   & \\     
   & max order & 6     & num units & 256  & & \\
   & std & 0.8     &   &  & & \\
   & $\lambda$ & 0.2    &   &   & & \\

   \cmidrule{1-3}
   \multirow{5}{*}{Bratley Sum < 500} & lr & 0.001   &    &  &    &  \\
   & num sums & 20  &  &                        &  & \\     
   & max order & 6     &   &  & & \\
   & std & 0.5     &   &  & & \\
   & $\lambda$ & 0.15    &   &   & & \\

   \cmidrule{1-3}
   \multirow{5}{*}{Bratley Sum >= 500} & lr & 0.001  &    &  &    &  \\
   & num sums & 75  &  &                       &  & \\     
   & max order & 6     &   &   & & \\
   & std & 0.9     &   &  & & \\
   & $\lambda$ & 0.12     &   &   & & \\
\end{tabular}
    \label{tab:hyperparams}
\end{table}

\begin{table}
   \caption{Model parameters.}
    \centering
    \begin{tabular}{cccccc}
experiment & model & num params & model size [MB] & max memory [MB] & train time [sec/iter] \\
\midrule
\multirow{3}{*}{Darcy flow} & DeepPCE & 43049558 & 164.456 & 4699.239 & 19.942 \\
& MLP & 20996098 & 80.094 & 1469.283 & 0.470 \\
& UNet & 31054163 & 118.462 & 4710.679 & 7.913 \\
\midrule
\multirow{3}{*}{Steady state diffusion} & DeepPCE & 11320558 & 43.419 & 2509.667 & 12.361 \\
 & MLP & 20996098 & 80.094 & 1472.880 & 0.437 \\
 & UNet & 7707471 & 29.402 & 4368.906 & 6.688 \\
\end{tabular}

    \label{tab:model_params}
\end{table}

\subsection{Main paper experiments: additional results}

\paragraph{Bratley Sum}
Relative MSEs for each of the 10 training runs for the Bratley Sum function with dimensions $D = [100, 250, 500, 750, 1000, 2000]$ for DeepPCE, $\text{PCE}_q$, $\text{PCE}_{OMP}$ and $\text{BA-PCE}_{OMP}$ are shown in \cref{tab:runs_bratleysum}.

\paragraph{PDEs}
Relative MSEs for each of the 10 training runs for the Dracy flow  and steady state diffusion experiments are shown in \cref{tab:runs_pdes}.

\begin{table}[h!]
    \caption{Relative MSEs for all 10 runs of the Bratley Sum experiment with dimensions $D = [100, 250, 500, 750, 1000, 2000]$. Experiments with traditional PCEs failed in higher dimensions due to memory constraints.}
    \vspace{\baselineskip}
    \centering
    \begin{tabular}{ccccc}
\toprule
\multicolumn{4}{c}{$D = 100$} \\
DeepPCE & $\text{PCE}_q$ & $\text{PCE}_{OMP}$ & $\text{BA-PCE}_{OMP}$ \\
\midrule
0.00037 & 0.12071 & 0.02520 & 0.00819 \\
0.00023 & 0.12683 & 0.02415 & 0.01159 \\
0.00069 & 0.11904 & 0.02398 & 0.00523 \\
0.00035 & 0.12808 & 0.02447 & 0.00993 \\
0.00025 & 0.12881 & 0.02521 & 0.00975 \\
0.00049 & 0.12326 & 0.02552 & 0.00984 \\
0.00059 & 0.11832 & 0.02441 & 0.00562 \\
0.00046 & 0.12064 & 0.02458 & 0.00878 \\
0.00024 & 0.12938 & 0.02364 & 0.00460 \\
0.00031 & 0.12412 & 0.02495 & 0.01170 \\
\bottomrule
\end{tabular}
\hspace{.3cm}
\begin{tabular}{ccccc}
\toprule
\multicolumn{4}{c}{$D = 250$} \\
DeepPCE & $\text{PCE}_q$ & $\text{PCE}_{OMP}$ & $\text{BA-PCE}_{OMP}$ \\
\midrule
0.00121 & 0.29169 & 0.01165 & 0.01320 \\
0.00076 & 0.29343 & 0.01169 & 0.01357 \\
0.00093 & 0.29658 & 0.01139 & 0.01418 \\
0.00144 & 0.29268 & 0.01267 & 0.01543 \\
0.00077 & 0.29097 & 0.01232 & 0.01457 \\
0.00105 & 0.29084 & 0.01305 & 0.01405 \\
0.00122 & 0.29580 & 0.01211 & 0.01424 \\
0.00169 & 0.28663 & 0.01290 & 0.01412 \\
0.00115 & 0.29051 & 0.01195 & 0.01352 \\
0.00104 & 0.29655 & 0.01226 & 0.01458 \\
\bottomrule
\end{tabular}
\begin{tabular}{cccc}
    & & & \\
\toprule
\multicolumn{3}{c}{$D = 500$} \\
DeepPCE & $\text{PCE}_{OMP}$ & $\text{BA-PCE}_{OMP}$ \\
\midrule
0.00210 & 0.01052 & 0.01053 \\
0.00159 & 0.01083 & 0.01134 \\
0.00284 & 0.01056 & 0.01050 \\
0.00673 & 0.01014 & 0.01100 \\
0.00197 & 0.01088 & 0.01143 \\
0.01389 & 0.01067 & 0.01146 \\
0.00291 & 0.00986 & 0.01084 \\
0.00391 & 0.01044 & 0.01120 \\
0.00575 & 0.01000 & 0.01107 \\
0.00353 & 0.01097 & 0.01170 \\
\bottomrule
\end{tabular}
\begin{tabular}{cc}
    & \\
\toprule
\multicolumn{2}{c}{$D = 750$} \\
DeepPCE & $\text{BA-PCE}_{OMP}$ \\
\midrule
0.34076 & 0.01121 \\
0.05260 & 0.01110 \\
0.00266 & 0.01202 \\
0.03848 & 0.01056 \\
0.00279 & 0.01110 \\
0.03562 & 0.01125 \\
0.10126 & 0.01136 \\
0.30875 & 0.01149 \\
0.10065 & 0.01097 \\
0.00103 & 0.01038 \\
\bottomrule
\end{tabular}
\begin{tabular}{c}
\\
\toprule
$D = 1000$ \\
DeepPCE \\
\midrule
0.02101 \\
0.00372 \\
0.01621 \\
0.15814 \\
0.00157 \\
0.04136 \\
0.03061 \\
0.00569 \\
0.03548 \\
0.07428 \\
\bottomrule
\end{tabular}
\begin{tabular}{cc}
    \\
\toprule
$D = 2000$ \\
DeepPCE \\
\midrule
0.04947 \\
0.05870 \\
0.08810 \\
0.15262 \\
0.14441 \\
0.16543 \\
0.12488 \\
0.04012 \\
0.13877 \\
0.11783 \\
\bottomrule
\end{tabular}

    \label{tab:runs_bratleysum}
\end{table}

\begin{table}[h!]
    \caption{Relative MSEs of DeepPCE, MLP and UNet on the Darcy flow dataset for all 10 test runs.}
    \vspace{\baselineskip}
    \centering
    \begin{tabular}{ccc}
\toprule
\multicolumn{3}{c}{Darcy flow} \\
DeepPCE & MLP & UNet \\
\midrule
0.002919 & 0.001632 & 0.000062 \\
0.003497 & 0.001668 & 0.000064 \\
0.003954 & 0.001698 & 0.000066 \\
0.005309 & 0.001885 & 0.000068 \\
0.005517 & 0.002159 & 0.000069 \\
0.006018 & 0.002181 & 0.000069 \\
0.006077 & 0.002353 & 0.000070 \\
0.006196 & 0.003312 & 0.000070 \\
0.007252 & 0.003597 & 0.000070 \\
0.018164 & 0.011898 & 0.000073 \\
\bottomrule
\end{tabular}
\hspace{1cm}
\begin{tabular}{ccc}
\toprule
\multicolumn{3}{c}{Steady state diffusion} \\
DeepPCE & MLP & UNet \\
\midrule
0.000754 & 0.000250 & 0.000074 \\
0.000813 & 0.000257 & 0.000076 \\
0.001164 & 0.000257 & 0.000078 \\
0.001186 & 0.000261 & 0.000079 \\
0.001376 & 0.000262 & 0.000080 \\
0.002412 & 0.000264 & 0.000080 \\
0.003373 & 0.000265 & 0.000081 \\
0.003886 & 0.000266 & 0.000083 \\
0.008011 & 0.000267 & 0.000087 \\
0.039037 & 0.000277 & 0.000090 \\
\bottomrule
\end{tabular}

    \label{tab:runs_pdes}
\end{table}

\newpage
\subsection{Additional experiments}

\subsubsection{PCE Benchmark}

\begin{figure}[h!]
    \centering
    \includegraphics[width=\textwidth]{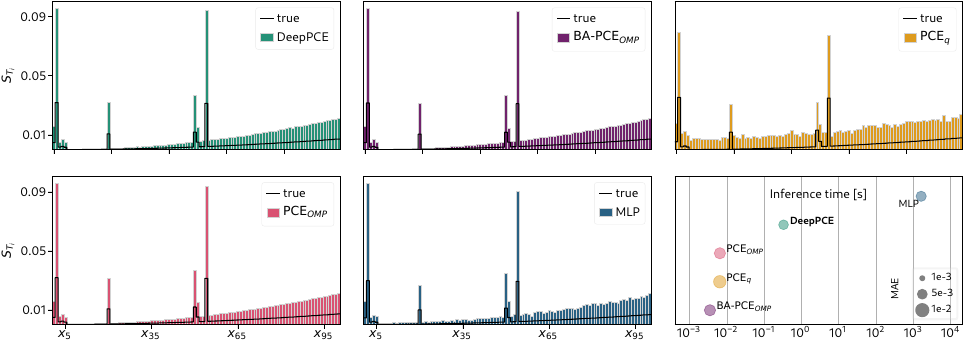}
        \caption{Total effect Sobol indices for the PCE benchmark dataset. 
        The DeepPCE and the sparse PCE variants ($\text{PCE}_{OMP}$, $\text{BA-PCE}_{OMP}$) are equally able to reproduce the true Sobol indices, while $\text{PCE}_q$ shows slightly worse performance. Using $10^8$ Monte Carlo samples, the Sobol indices obtained from the MLP are comparable to the results of the other models, but exhibiting high computational cost.}
    \label{fig:sobol_pcebenchmark}
\end{figure}

We additionally computed total effect Sobol indices for the PCE benchmark function (\cref{eq:pce_benchmark}), comparing the DeepPCE with the other PCE variants ($\text{PCE}_q$, $\text{PCE}_{OMP}$, $\text{BA-PCE}_{OMP}$) and Monte Carlo-approximated Sobol indices of an MLP, using $10^8$ samples.
The results are shown in \cref{fig:sobol_pcebenchmark}. 
While all models slightly overestimate the variance contributions of each variable, the DeepPCE and the sparse PCE variants ($\text{PCE}_{OMP}$, $\text{BA-PCE}_{OMP}$) manage to capture the general pattern of the total effect Sobol indices. 
The PCE using a simple hyperbolic truncation ($\text{PCE}_{q}$) shows the worst performance of all PCEs. 
The accuracy of the MLP Sobol indices is comparable to that of the other models, but the computation time is prohibitive.

\section{ABLATION STUDIES}

We performed ablation studies to assess convergence performance with different scope sizes on the Bratley dataset with $D = 100$ and the steady state diffusion PDE, using scope sizes $S = [1, 2, 3]$ while keeping the other hyperparameters fixed according to \cref{tab:hyperparams}.
Convergence rates and predictive performance deteriorates with larger scope sizes, especially for the larger steady state diffusion dataset.
A likely reason is the presence of higher rank interaction terms in the expansion, amplifying the problem of vanishing/exploding values in the circuit.
Additionally, the initialization is based on the variance scaling scheme shown in \cref{sec:training}, using a decay hyperparameter $\lambda$ tuned specifically for a scope size of 1.
Changing the scope sizes at the input layer thus requires retuning of $\lambda$.

\begin{figure}[ht!]
    \centering
    \begin{minipage}[t]{0.47\textwidth}
        \centering
      \includegraphics[width=.8\linewidth]{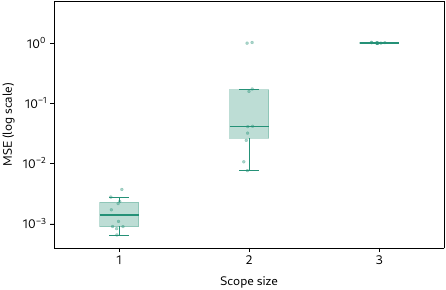}
    \end{minipage}
    \hfill{}
    \begin{minipage}[t]{0.47\textwidth}
      \includegraphics[width=.8\linewidth]{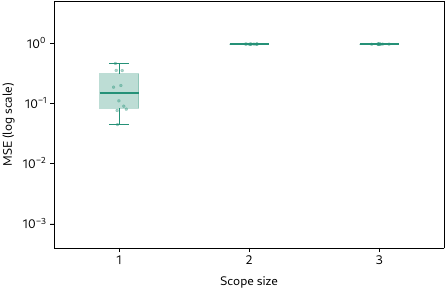}
    \end{minipage}
        \caption{Test performance from 10 runs for a DeepPCE with different scope size parameter for the Bratyel function with $D = 100$ (left) and the steady-state diffusion dataset (right).
        Results show that the scope size has a strong effect on performance and convergence rates of the DeepPCE.}
    \label{fig:ablation}
\end{figure}

\end{document}